\documentclass[11pt, a4paper, goog]{google}

\usepackage[authoryear, sort&compress, round]{natbib}
\usepackage[T1]{fontenc}
\usepackage{lmodern}
\usepackage[scale=0.92]{tgheros}
\usepackage{microtype}
\usepackage{amsmath,amssymb}
\usepackage[utf8]{inputenc}
\usepackage{booktabs}
\usepackage{multirow}
\usepackage{longtable}
\usepackage{makecell}
\usepackage{array}
\usepackage{graphicx}
\usepackage{xcolor}
\usepackage{colortbl}
\usepackage{pifont}
\usepackage{tikz}
\usetikzlibrary{arrows.meta,positioning,fit,backgrounds,shapes.geometric,shadows,calc}
\usepackage{tikzpeople}
\definecolor{plePrint}{HTML}{F5F7FA}
\definecolor{pleSheet}{HTML}{FFFFFF}
\definecolor{pleWash}{HTML}{E8ECF6}
\definecolor{pleHairline}{HTML}{C7CFE3}
\definecolor{pleBorder}{HTML}{C8D2E0}
\definecolor{pleBlueline}{HTML}{1F2F86}
\definecolor{pleRedline}{HTML}{E56B4E}
\colorlet{pleCoFill}{pleRedline}
\colorlet{pleCoText}{pleRedline}
\definecolor{pleGraphite}{HTML}{5D6577}
\definecolor{pleInk}{HTML}{0B192C}
\definecolor{plePillBg}{HTML}{EBEEF7}
\definecolor{plePillBdr}{HTML}{B7BEDB}
\definecolor{pleRowAlt}{HTML}{F7F9FC}
\definecolor{infraBlue}{HTML}{1F2F86}
\definecolor{infraOrange}{HTML}{E56B4E}
\definecolor{allGreen}{HTML}{E8ECF6} 
\arrayrulecolor{pleHairline}

\newsavebox{\plesheetbox}
\newlength{\plesheetw}
\newlength{\plesheeth}

\usepackage{pgfplots}
\pgfplotsset{compat=1.17}
\usepackage{caption}
\usepackage{subcaption}
\usepackage{enumitem}
\usepackage{xcolor}
\usepackage{placeins}

\usepackage[colorlinks=true]{hyperref}   
\hypersetup{
  linkcolor=pleBlueline,
  citecolor=pleBlueline,
  urlcolor=pleBlueline
}

\usepackage{cleveref}   

\newcommand{\bench}{PLE}

\newif\ifshownotes\shownotestrue

\newcommand{\modelset}{Claude Opus~5.5, Claude Opus~5, Claude Sonnet~5.5, Claude Fable~5, GPT~6~Astra, Gemini~4~Argon, Gemini~3.8~Flash, Gemini~3.1~Pro, Grok~4.7, GLM~5.2, Qwen3~Coder~480B, MiniMax~M2, Kimi~K2~Thinking and Llama~4~Maverick}

\uselogo{} 

\title{\bench{}: Physical-Engineering's Last Exam}

\correspondingauthor{rujunh@google.com, yufanzhuang@google.com, chenyulee@google.com}

\renewcommand{\today}{}

\author[1]{Rujun Han}
\author[1]{Yufan Zhuang}
\author[1]{Arijit Ray}
\author[1]{Han Yu}
\author[1]{Zhongying CuiZhu}
\author[1]{George Lee}
\author[1]{Yifei Ming}
\author[2]{Binz Roy}
\author[2]{Varun Sakalkar}
\author[1]{Vishy Tirumalashetty}
\author[1]{Burak Gokturk}
\author[1]{Tomas Pfister}
\author[1]{Chen-Yu Lee}

\affil[1]{Google Cloud AI Research}
\affil[2]{Google}

\begin{abstract}
Frontier language models now score above 90\% on popular knowledge benchmarks and are
increasingly deployed as agents that carry out multi-hour professional work. Yet no existing
benchmark tests whether they can perform the real-world, economically valuable work of a
\emph{professional physical engineer}, such as modelling physics and legal code requirements across a multi-format project corpus, optimising material and installation costs, and issuing 2D/3D engineering
design packages that a licensed reviewer would sign off on. Existing engineering benchmarks focus on
textbook question answering, isolated tool calls or self-contained toy design problems. As a result, most of these benchmarks are already saturated by leading agents and solving them is far away from delivering real engineering design. We
introduce \bench{}, a benchmark of 85 long-horizon, agentic tasks in professional
physical engineering consisting of electrical, civil, structural and mechanical engineering, where solving
a task means completing a realistic, billable engineering deliverable. \bench{} combines
two complementary curation paths. \emph{Expert-Authored} tasks are created end to end by practising
engineers from a professional data provider's expert pool.
\emph{Co-Authored} tasks start from problem statements written by practising engineers; an
adversarial Creator--Solver agent pipeline constructs each task's realistic project environment, executable grader and reference deliverables before sending them to expert engineers for validation and difficulty intensification. We evaluate a variety of frontier LLMs
under seven publicly available agent harnesses. Even the strongest configuration reaches a mean score of only 50.6\%, while spending 27M tokens and
63 minutes per task on average, demonstrating the substantial gap between current
agents and autonomous physical engineering design.
We will release
a public set of 29 of the 85 tasks, together with their graders and the agent harness, to help the community track progress toward
economically valuable AI physical engineers, and hold the remaining tasks private for an open competition and leaderboard.

\vspace{4pt}
{\centering
\raisebox{-0.16\height}{\includegraphics[height=1.05em]{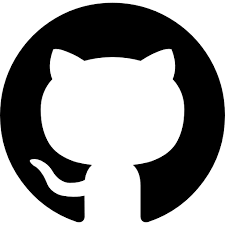}}\hspace{0.32em}\href{https://github.com/google-research/ple}{\texttt{github.com/google-research/ple}}%
\hspace{1.15em}\textcolor{gray!45}{\rule[0.02em]{0.6pt}{0.9em}}\hspace{1.15em}%
\raisebox{-0.16\height}{\includegraphics[height=1.05em]{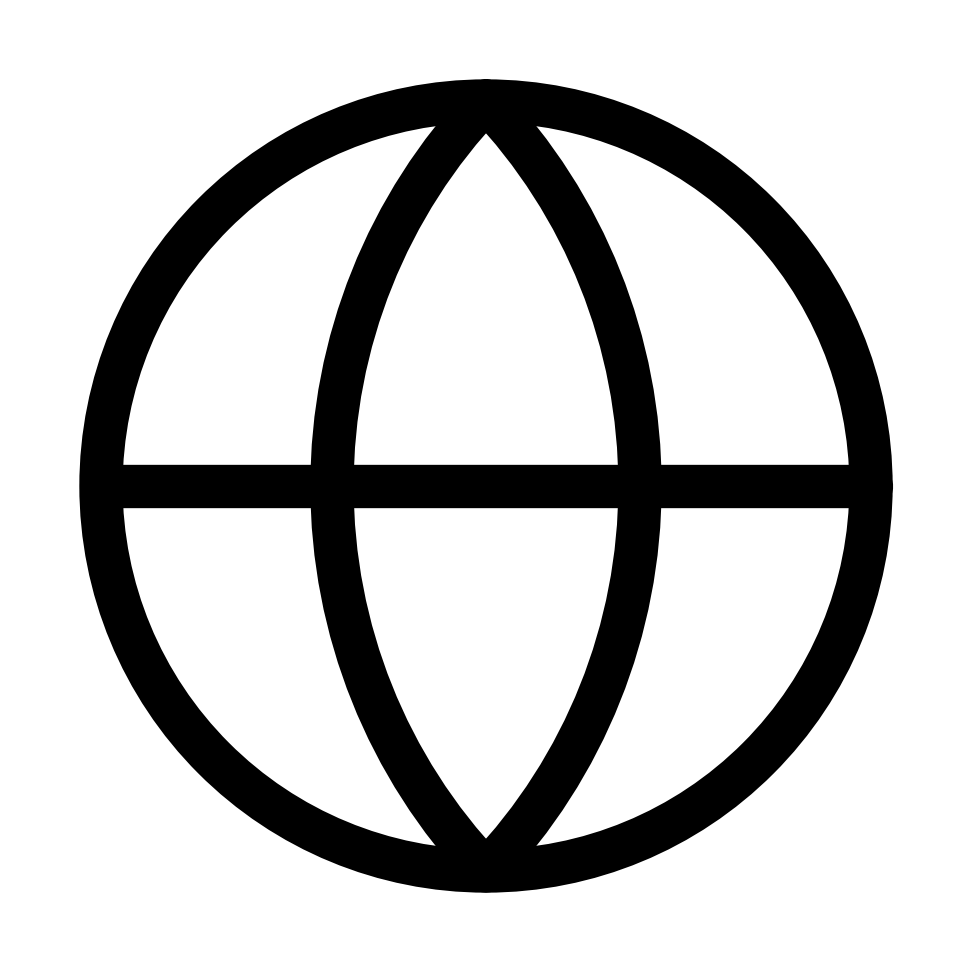}}\hspace{0.32em}\href{https://ple-bench.com/}{\texttt{ple-bench.com}}%
\par}
\end{abstract}

\begin{document}

\maketitle

\section{Introduction}
\label{sec:intro}

\begin{figure*}[t]
\centering
\setlength{\tabcolsep}{3pt}%
\renewcommand{\arraystretch}{0}%
\begin{tabular}{@{}cc@{}}
  \includegraphics[width=0.491\textwidth]{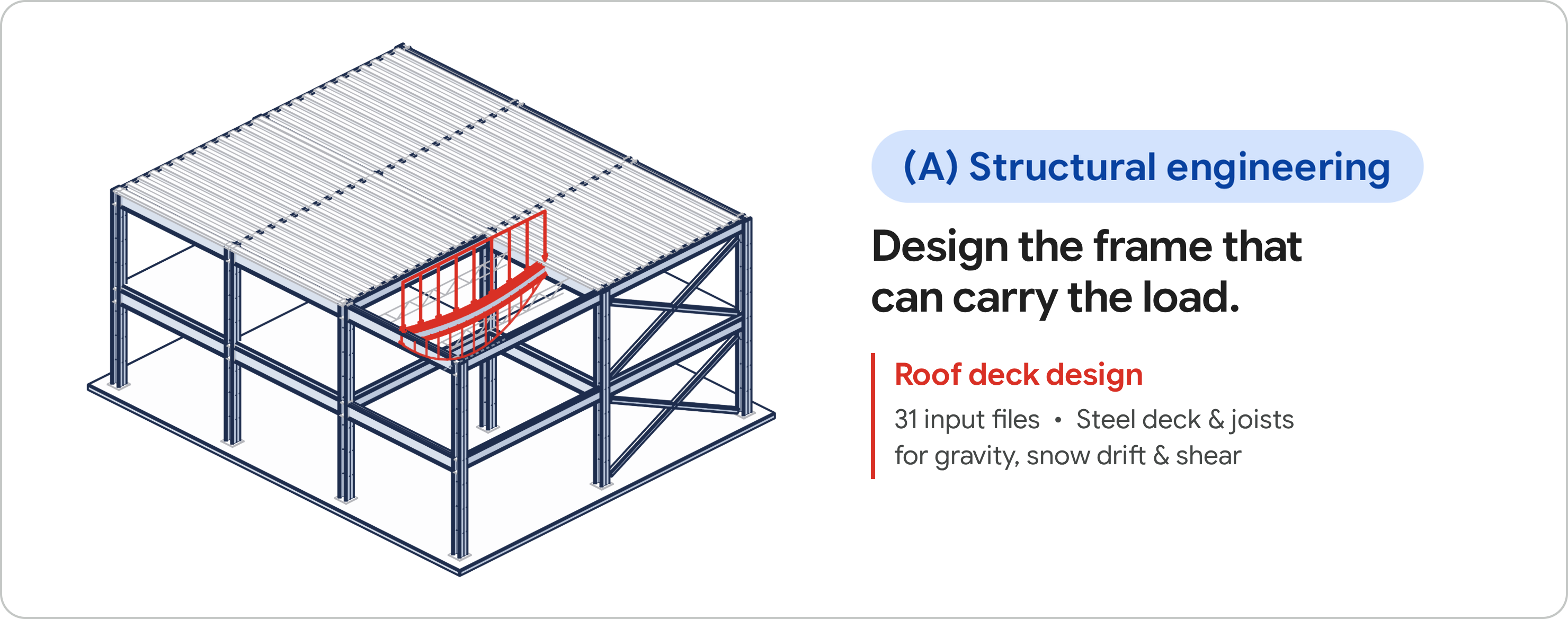} &
  \includegraphics[width=0.491\textwidth]{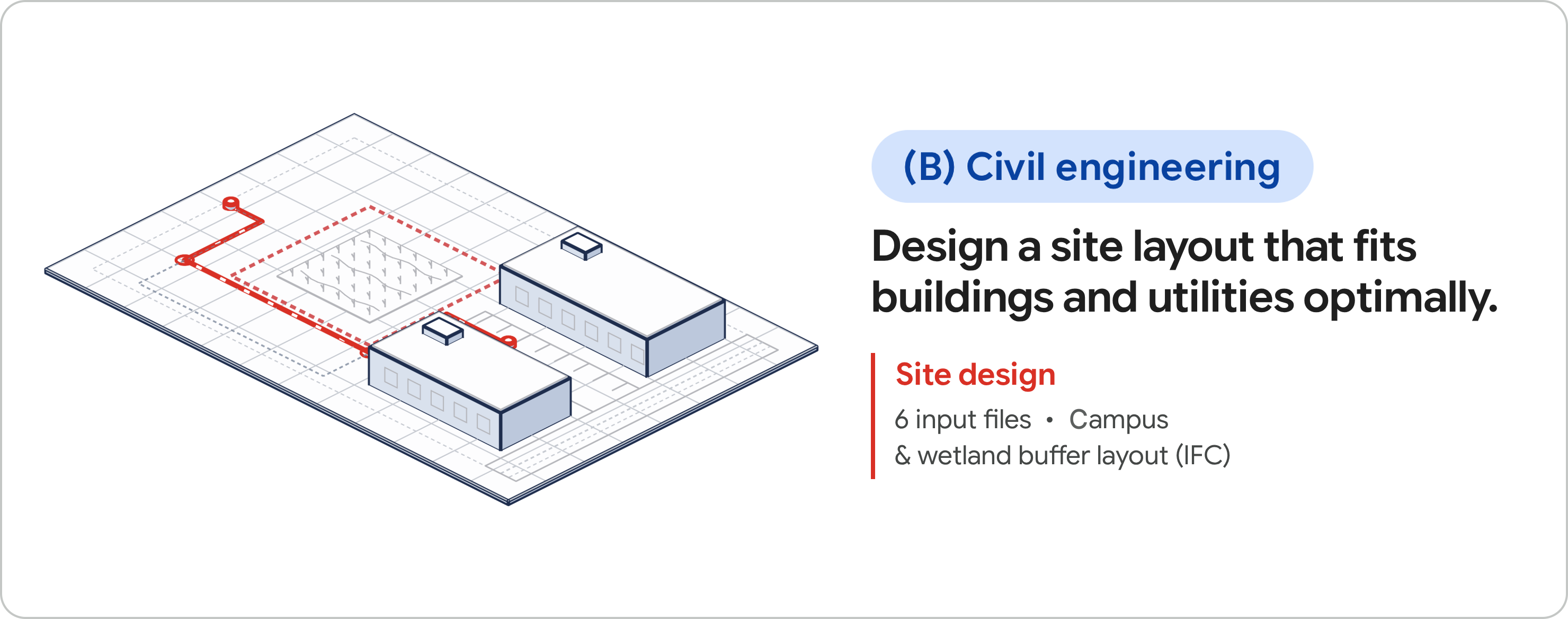} \\[5pt]
  \includegraphics[width=0.491\textwidth]{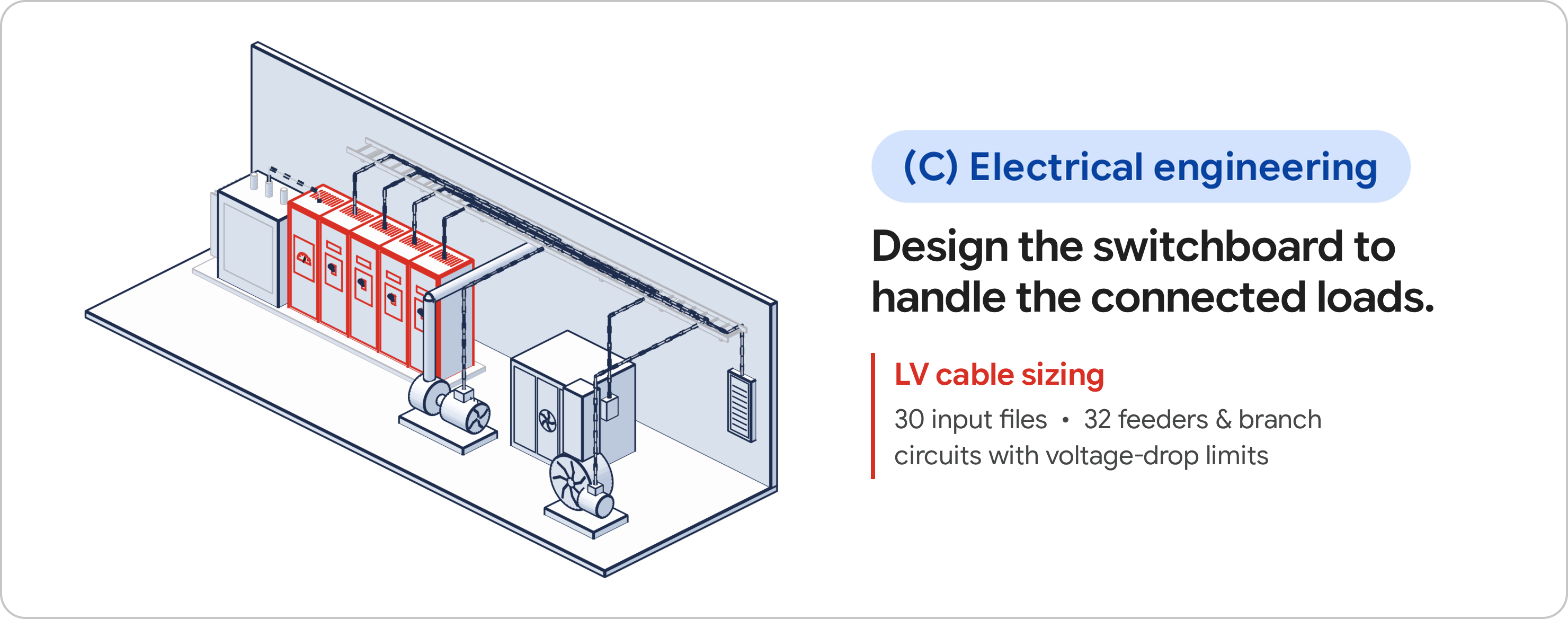} &
  \includegraphics[width=0.491\textwidth]{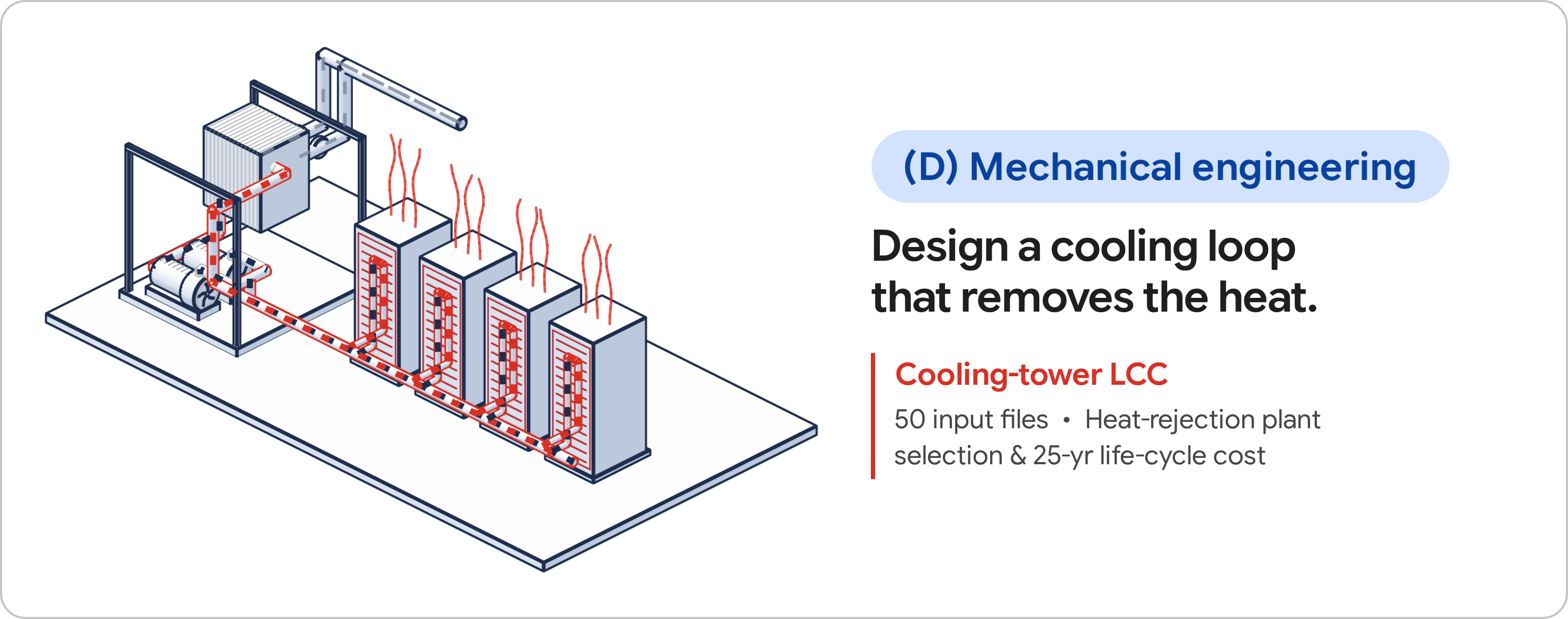}
\end{tabular}
\vspace{-2pt}
\caption{Overview of the four professional physical-engineering domains in \bench{}: \textbf{(A)}~Structural, \textbf{(B)}~Civil, \textbf{(C)}~Electrical, and \textbf{(D)}~Mechanical Engineering, illustrated with representative tasks.}
\label{fig:domain_overview}
\end{figure*}

The physical systems and infrastructure that modern society runs on (hospitals, schools and
university campuses, office and laboratory buildings, power distribution, water and transportation
networks) are designed by \emph{professional physical engineers}. Demand for this engineering work
is rising rapidly: the American Society of Civil Engineers estimates a \$3.7 trillion gap over
2024--2033 between what U.S.\ physical infrastructure needs to reach a state of good repair and
forecasted investment~\citep{asce2025reportcard}, and worldwide infrastructure investment needs are
forecast to reach \$94 trillion by 2040~\citep{gihub2017outlook}.
Delivering even a single campus building, such as a new university laboratory, takes electrical,
mechanical, structural and civil engineers to size its power distribution and feeders, its heating,
cooling and ventilation plant, its frame and foundations, and its site grading and stormwater
systems. They work to legally binding codes and physical laws (\cref{fig:domain_overview}), and
their designs are signed off by licensed professionals, because
errors in the design are costly and potentially catastrophic. Engineering design services represent hundreds
of billions of dollars in annual economic activity (U.S.\ engineering-services firms alone report
over \$300 billion in yearly revenue~\citep{census2022sas}), making professional physical engineering a
high-value frontier for AI agents.

As \cref{fig:domain_overview} illustrates across the four domains of \bench{}, a real-world physical engineering job requires far more than answering a textbook prompt:
\textbf{(A)~Structural Engineering} (\emph{Roof deck design}) requires sizing steel roof decking, open-web joists and lateral diaphragm framing under gravity, snow-drift and shear loads so member demand stays within code capacity;
\textbf{(B)~Civil Engineering} (\emph{Site design}) requires laying out a phased campus, access roads and utility corridors around delineated wetland buffers as a 3D IFC model;
\textbf{(C)~Electrical Engineering} (\emph{LV cable sizing}) requires sizing low-voltage switchboards, overhead cable trays and 32~feeder and branch circuits under ampacity and voltage-drop limits; and
\textbf{(D)~Mechanical Engineering} (\emph{Cooling-tower LCC}) requires selecting heat-rejection equipment, pumps and piping to meet thermal load within hydraulic flow limits while minimising 25-year life-cycle cost.
Each task is long-horizon, spread across many documents in
native formats, and combines retrieval, spatial and physical modelling, code compliance and cost
optimisation, yet it is unusually measurable. Its deliverables are real engineering work products that can be checked
against the governing documents and physics, so both the eligibility and the economic quality of a design can be scored objectively.

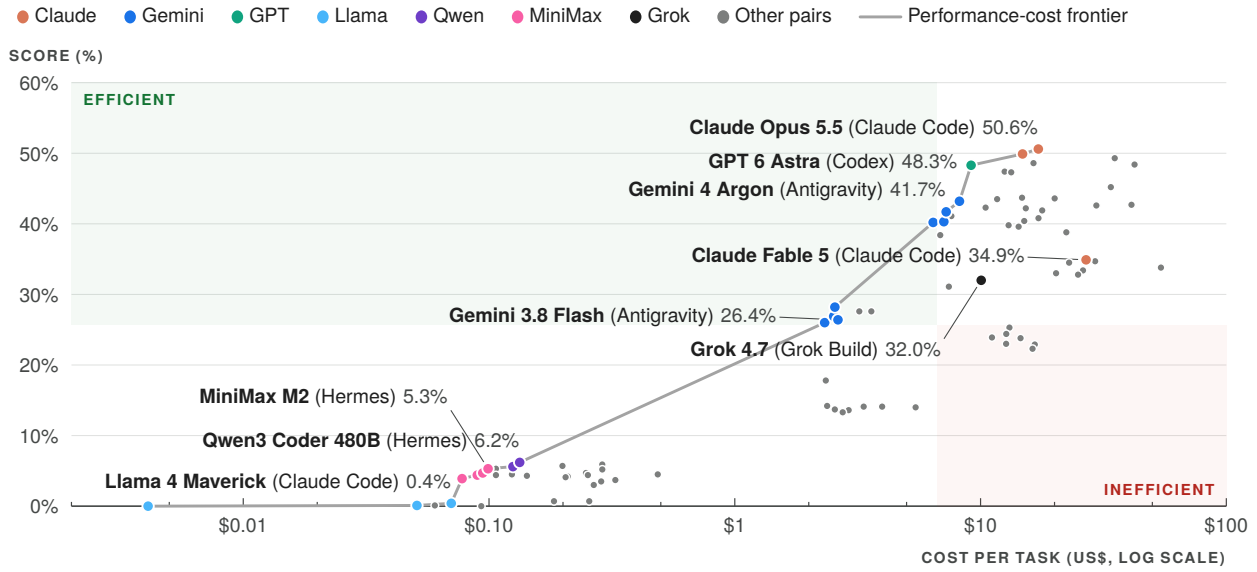
\begin{figure}[t]
\centering
\definecolor{famAnthropic}{HTML}{D97757}%
\definecolor{famGoogle}{HTML}{1A73E8}%
\definecolor{famOpenai}{HTML}{10A37F}%
\definecolor{famMeta}{HTML}{48B5FA}%
\definecolor{famAlibaba}{HTML}{6E3EC7}%
\definecolor{famDeepseek}{HTML}{A088FF}%
\definecolor{famMinimax}{HTML}{F85EA5}%
\definecolor{famOther}{HTML}{1F1F1F}%
\definecolor{webInk}{HTML}{1F1F1F}%
\definecolor{webGraphite}{HTML}{444746}%
\definecolor{webHairline}{HTML}{C4C7C5}%
\definecolor{webQuiet}{HTML}{7C7E7D}%
\definecolor{webLine}{HTML}{A3A3A3}%
\definecolor{webGreen}{HTML}{1E8E3E}%
\definecolor{webGreenText}{HTML}{137333}%
\definecolor{webRed}{HTML}{D93025}%
\definecolor{webRedText}{HTML}{B3261E}%
\begin{tikzpicture}
\begin{axis}[
  scale only axis, width=0.92\linewidth, height=5.6cm,
  xmode=log, log basis x=10, xmin=0.002, xmax=100, ymin=0, ymax=0.6,
  axis lines*=left, axis on top=false, clip=false,
  x axis line style={draw=webGraphite, line width=0.6pt, -}, y axis line style={draw=none},
  xtick={0.01,0.1,1,10,100}, xticklabels={\$0.01,\$0.10,\$1,\$10,\$100},
  minor xtick={0.002,0.005,0.02,0.05,0.2,0.5,2,5,20,50},
  xtick style={draw=webGraphite, line width=0.45pt}, major tick length=3pt, minor tick length=1.6pt,
  ytick={0.0,0.1,0.2,0.3,0.4,0.5,0.6}, yticklabels={0\%,10\%,20\%,30\%,40\%,50\%,60\%},
  ytick style={draw=none}, ymajorgrids, major grid style={draw=webHairline!55, line width=0.35pt},
  tick label style={font=\scriptsize\sffamily, text=webGraphite}, yticklabel style={xshift=-1pt},
  legend style={font=\scriptsize\sffamily, text=webInk, draw=none, fill=none, legend columns=-1,
    at={(0,1)}, anchor=south west, yshift=17pt, xshift=-24pt,
    /tikz/every even column/.append style={column sep=8pt}},
  legend cell align=left, legend image post style={xscale=0.9, mark size=2.8pt},
]
\fill[webGreen, opacity=0.05] (axis cs:0.002,0.2565) rectangle (axis cs:6.63455,0.6);
\fill[webRed, opacity=0.045] (axis cs:6.63455,0) rectangle (axis cs:100,0.2565);
\node[font=\fontsize{6.2}{7}\selectfont\sffamily\bfseries, text=webGreenText, anchor=north west, inner sep=0pt] at ([shift={(4pt,-4pt)}]axis cs:0.002,0.6) {\textls[110]{EFFICIENT}};
\node[font=\fontsize{6.2}{7}\selectfont\sffamily\bfseries, text=webRedText, anchor=south east, inner sep=0pt] at ([shift={(-4pt,4pt)}]axis cs:100,0) {\textls[110]{INEFFICIENT}};
\addplot[draw=webLine, line width=1.1pt, line join=round, line cap=round, forget plot] coordinates {
  (0.0041,0.0) (0.0509,0.001) (0.0702,0.004) (0.0777,0.039) (0.0896,0.044) (0.0943,0.047) (0.0991,0.053) (0.1251,0.056) (0.1333,0.062) (2.3213,0.26) (2.5258,0.269) (2.5539,0.282) (6.4139,0.402) (7.0905,0.403) (7.2445,0.417) (8.207,0.432) (9.1513,0.483) (14.8198,0.499) (17.1839,0.506)
};
\addplot[only marks, mark=*, mark size=1.45pt, draw=white, line width=0.55pt, fill=webQuiet, forget plot] coordinates {
  (35.1491,0.493) (16.4225,0.486) (42.3236,0.484) (12.5302,0.474) (13.3282,0.473) (33.8914,0.452) (14.7483,0.437) (20.0202,0.436) (11.6804,0.435) (41.122,0.427) (29.5915,0.426) (10.4834,0.423) (15.2789,0.422) (17.8169,0.419) (7.6092,0.411) (17.2086,0.408) (15.0557,0.404) (13.0036,0.398) (14.2859,0.396) (22.3343,0.388) (6.8552,0.384) (29.2532,0.347) (22.9241,0.345) (54.1939,0.338) (26.1095,0.334) (20.29,0.33) (24.9302,0.328) (7.4318,0.311) (3.5946,0.276) (3.2109,0.276) (13.1067,0.253) (12.7227,0.244) (11.1281,0.239) (14.5452,0.238) (12.7064,0.23) (16.6189,0.229) (16.2752,0.223) (2.3457,0.178) (2.3756,0.142) (3.9778,0.141) (3.3456,0.141) (5.4424,0.14) (2.553,0.137) (2.9057,0.136) (2.7485,0.133) (0.2889,0.059) (0.199,0.057) (0.1067,0.053) (0.289,0.052) (0.248,0.047) (0.4859,0.045) (0.1239,0.045) (0.1067,0.044) (0.2525,0.044) (0.1426,0.043) (0.209,0.042) (0.2049,0.041) (0.3266,0.037) (0.286,0.035) (0.2668,0.03) (0.1838,0.007) (0.2557,0.007) (0.0602,0.001) (0.093,0.0)
};
\addplot[only marks, mark=*, mark size=2.15pt, draw=white, line width=0.7pt, fill=famAnthropic] coordinates { (14.8198,0.499) (17.1839,0.506) (26.8697,0.349) };
\addlegendentry{Claude}
\addplot[only marks, mark=*, mark size=2.15pt, draw=white, line width=0.7pt, fill=famGoogle] coordinates { (2.3213,0.26) (2.5258,0.269) (2.5539,0.282) (6.4139,0.402) (7.0905,0.403) (7.2445,0.417) (8.207,0.432) (2.6293,0.264) };
\addlegendentry{Gemini}
\addplot[only marks, mark=*, mark size=2.15pt, draw=white, line width=0.7pt, fill=famOpenai] coordinates { (9.1513,0.483) };
\addlegendentry{GPT}
\addplot[only marks, mark=*, mark size=2.15pt, draw=white, line width=0.7pt, fill=famMeta] coordinates { (0.0041,0.0) (0.0509,0.001) (0.0702,0.004) };
\addlegendentry{Llama}
\addplot[only marks, mark=*, mark size=2.15pt, draw=white, line width=0.7pt, fill=famAlibaba] coordinates { (0.1251,0.056) (0.1333,0.062) };
\addlegendentry{Qwen}
\addplot[only marks, mark=*, mark size=2.15pt, draw=white, line width=0.7pt, fill=famMinimax] coordinates { (0.0777,0.039) (0.0896,0.044) (0.0943,0.047) (0.0991,0.053) };
\addlegendentry{MiniMax}
\addplot[only marks, mark=*, mark size=2.15pt, draw=white, line width=0.7pt, fill=famOther] coordinates { (10.0612,0.32) };
\addlegendentry{Grok}
\addlegendimage{only marks, mark=*, mark size=1.45pt, draw=white, line width=0.55pt, fill=webQuiet}
\addlegendentry{Other pairs}
\addlegendimage{draw=webLine, line width=1.1pt, line cap=round}
\addlegendentry{Performance-cost frontier}
\node[font=\scriptsize\sffamily, text=webInk, anchor=south east, xshift=1pt, yshift=3pt, inner sep=1pt] (lbl0) at (axis cs:17.1839,0.506) {\textbf{Claude Opus 5.5} (Claude Code)\hspace{0.35em}\textcolor{webGraphite}{50.6\%}};
\node[font=\scriptsize\sffamily, text=webInk, anchor=east, xshift=-4pt, yshift=1pt, inner sep=1pt] (lbl1) at (axis cs:9.1513,0.483) {\textbf{GPT 6 Astra} (Codex)\hspace{0.35em}\textcolor{webGraphite}{48.3\%}};
\node[font=\scriptsize\sffamily, text=webInk, anchor=south east, xshift=1pt, yshift=3pt, inner sep=1pt] (lbl2) at (axis cs:7.2445,0.417) {\textbf{Gemini 4 Argon} (Antigravity)\hspace{0.35em}\textcolor{webGraphite}{41.7\%}};
\node[font=\scriptsize\sffamily, text=webInk, anchor=east, xshift=-22pt, yshift=1pt, inner sep=1pt] (lbl3) at (axis cs:26.8697,0.349) {\textbf{Claude Fable 5} (Claude Code)\hspace{0.35em}\textcolor{webGraphite}{34.9\%}};
\draw[webGraphite, line width=0.4pt, shorten >=3pt] (lbl3.east) -- (axis cs:26.8697,0.349);
\node[font=\scriptsize\sffamily, text=webInk, anchor=north east, xshift=-14pt, yshift=-22pt, inner sep=1pt] (lbl4) at (axis cs:10.0612,0.32) {\textbf{Grok 4.7} (Grok Build)\hspace{0.35em}\textcolor{webGraphite}{32.0\%}};
\draw[webGraphite, line width=0.4pt, shorten >=3pt] (lbl4.north east) -- (axis cs:10.0612,0.32);
\node[font=\scriptsize\sffamily, text=webInk, anchor=east, xshift=-22pt, yshift=1pt, inner sep=1pt] (lbl5) at (axis cs:2.6293,0.264) {\textbf{Gemini 3.8 Flash} (Antigravity)\hspace{0.35em}\textcolor{webGraphite}{26.4\%}};
\draw[webGraphite, line width=0.4pt, shorten >=3pt] (lbl5.east) -- (axis cs:2.6293,0.264);
\node[font=\scriptsize\sffamily, text=webInk, anchor=south east, xshift=1pt, yshift=3pt, inner sep=1pt] (lbl6) at (axis cs:0.1333,0.062) {\textbf{Qwen3 Coder 480B} (Hermes)\hspace{0.35em}\textcolor{webGraphite}{6.2\%}};
\node[font=\scriptsize\sffamily, text=webInk, anchor=south east, xshift=-14pt, yshift=22pt, inner sep=1pt] (lbl7) at (axis cs:0.0991,0.053) {\textbf{MiniMax M2} (Hermes)\hspace{0.35em}\textcolor{webGraphite}{5.3\%}};
\draw[webGraphite, line width=0.4pt, shorten >=3pt] (lbl7.south east) -- (axis cs:0.0991,0.053);
\node[font=\scriptsize\sffamily, text=webInk, anchor=south east, xshift=1pt, yshift=3pt, inner sep=1pt] (lbl8) at (axis cs:0.0702,0.004) {\textbf{Llama 4 Maverick} (Claude Code)\hspace{0.35em}\textcolor{webGraphite}{0.4\%}};
\node[font=\fontsize{6.2}{7}\selectfont\sffamily\bfseries, text=webGraphite, anchor=south west, inner sep=0pt] at ([shift={(-24pt,7pt)}]rel axis cs:0,1) {\textls[110]{SCORE (\%)}};
\node[font=\fontsize{6.2}{7}\selectfont\sffamily\bfseries, text=webGraphite, anchor=north east, inner sep=0pt] at ([yshift=-17pt]rel axis cs:1,0) {\textls[110]{COST PER TASK (US\$, LOG SCALE)}};
\end{axis}
\end{tikzpicture}
\caption{Score and cost of every model (harness) pair on the 85 tasks of \bench{} (mean of three runs; cost per task at list prices, log scale). Colored dots joined by the line form the performance-cost frontier (the highest score for the cost), colored by model family; Claude Fable~5 in Claude Code, Grok~4.7 in Grok Build and Gemini~3.8~Flash in Antigravity are also shown in color off the frontier; gray dots are the remaining pairs. Shading splits the chart at the median cost (\$6.63) and median score (25.7\%): more score for less cost at upper left, less score for more cost at lower right. The best configuration, Claude Opus~5.5 in Claude Code, scores 50.6\% at \$17.18 per task.}
\label{fig:pareto}
\end{figure}

No existing benchmark measures this economically valuable work comprehensively, and the benchmarks that come closest are
already too easy for leading agents. Existing engineering benchmarks for language models are mostly question answering over
textbook problems or rule documents~\citep{zhou2025engibench,doris2024designqa,zhou2024elecbench},
or self-contained toy design and optimisation problems whose full specification fits in the
prompt~\citep{guo2025engdesign,chi2026frontiereng}. Others target an isolated tool or modality, such
as finite-element software~\citep{mudur2025feabench}, CAD generation~\citep{khan2024text2cad,doris2026cadbench},
circuit design~\citep{lai2024analogcoder,bhandari2024masalachai} or drawing revision~\citep{li2025drafterbench}.
Solving those benchmarks does not get real engineering design work done, i.e., none requires an agent to navigate a realistic, messy, multi-format project corpus and issue a complete, billable engineering deliverable. The general professional-work benchmarks
GDPval~\citep{patwardhan2025gdpval} and APEX-Agents~\citep{vidgen2026apexagents} are closest in
spirit. They grade with expert pairwise preferences or expert rubrics. However, in physical engineering, a rubric of
binary checks alone cannot tell a merely code-compliant design from an economically competitive one, e.g. two
wire schedules can both satisfy every code check while one costs over 20\% more to install.

We introduce \bench{}, a benchmark of 85 expert-grade professional physical engineering tasks for
AI agents, spanning electrical, civil, structural and mechanical engineering. Unlike prior academic or single-tool benchmarks, \bench{} is built for realism and direct economic value: solving a task in \bench{} means completing a real-world engineering assignment. Each task places an agent in a sandboxed
environment that holds a realistic project file system of PDFs, spreadsheets, CAD/BIM drawings (DXF/DWG/IFC),
correspondence and logs. The agent has tool access for documents, spreadsheets and code execution,
and must produce cited, production-format deliverable files such as engineering memos, calculation workbooks, equipment schedules or 2D/3D drawings. We
build the benchmark through two complementary routes:
\begin{itemize}[leftmargin=*,itemsep=2pt]
  \item \textbf{Expert-Authored tasks.} Practising engineers from the expert pool of our data vendor (Mercor) create
  tasks end to end inside realistic project ``environments'', written as the real engineering requests a colleague or client would
  send on an active job. Each task is graded by a weighted rubric of binary, evidence-bound
  criteria~\citep{vidgen2026apexagents} (\cref{sec:expert}).
  \item \textbf{Co-Authored tasks.} Practising engineers write the task briefs, and an adversarial
  Creator--Solver agent pipeline builds each task's environment and grader from real, licensed engineering
  documents~\citep{xu2026sage}. Every task is iteratively
  redesigned until a frontier solver agent no longer saturates it, and passes an independent
  ground-truth audit. Mercor's experts then validate or intensify the difficulty of every task (\cref{sec:coauth}).
  These tasks are scored by both rubric-style constraints and an economic or performance
  objective, which complements the
  rubric-only scoring of the expert-authored tasks.
\end{itemize}
We evaluate 14 frontier models under seven publicly available agent harnesses (\cref{fig:pareto}). The best
configuration, Claude Opus~5.5 in Claude Code, reaches a mean score of only 50.6\%,
which demonstrates the difficulty of real-world physical engineering work.

In summary, our contributions are:
\begin{enumerate}[leftmargin=*,itemsep=2pt]
  \item \textbf{\bench{}}, the first agentic benchmark for realistic, economically valuable professional physical engineering
  work. It comprises 85 tasks across electrical, civil, structural and mechanical engineering, each grounded in a multi-format project corpus with a mean of
  61 files.
  \item \textbf{A two-route curation methodology} pairing expert authoring with human-agent co-authoring to scale up task creation and create objective-oriented, challenging tasks.
  \item \textbf{Comprehensive baseline results} for 14 frontier models across seven agent harnesses, analysing per-discipline difficulty, run-to-run variance, performance-cost trade-offs, etc.
  \item \textbf{An open competition and leaderboard} for AI physical-engineering agents. We evaluate submitted agent implementations on \bench{}, posting results to a public leaderboard.
\end{enumerate}

\section{Related Work}
\label{sec:related}

\begin{table}[t]
\centering
\caption{Comparison with representative benchmarks. \emph{\#Physical Engineering}: tasks in
professional physical engineering (electrical, civil, structural and mechanical engineering).
\emph{Real Work Tasks}: tasks reflecting authentic professional workflows that get real
job work done, rather than textbook problems, competitions, or isolated sub-tasks.
\emph{Expert Curation}: tasks authored or validated by domain professionals.
\emph{Large Input Corpus}: the agent must navigate a realistic multi-format document set
(e.g.\ PDF, spreadsheet, CAD). \emph{Large Output Artifacts}: the output is a production
work-product file rather than an answer string. \emph{Objective Function}: programmatic
scoring of solution quality or correctness. \emph{Rubric}: expert rubric or expert-judgment grading.}
\label{tab:related}
\definecolor{webInk}{HTML}{1F1F1F}%
\definecolor{webGraphite}{HTML}{444746}%
\definecolor{webHairline}{HTML}{C4C7C5}%
\definecolor{webWash}{HTML}{F0F4F9}%
\definecolor{webBlue}{HTML}{0B57D0}%
\definecolor{webNo}{HTML}{8E918F}%
\colorlet{webTint}{webBlue!13}%
\footnotesize\sffamily\color{webInk}%
\newsavebox{\relYes}\newsavebox{\relOwn}\newsavebox{\relNo}%
\sbox{\relYes}{\tikz[baseline=0pt]\draw[webBlue, line width=1.05pt, line cap=round, line join=round]
  (0,2.9pt) -- (2.5pt,0.5pt) -- (7.2pt,5.4pt);}%
\sbox{\relOwn}{\tikz[baseline=0pt]{\fill[webBlue] (3.6pt,2.9pt) circle (4.7pt);
  \draw[white, line width=1.0pt, line cap=round, line join=round] (1.5pt,3.0pt) -- (3.0pt,1.5pt) -- (5.9pt,4.5pt);}}%
\sbox{\relNo}{\tikz[baseline=0pt]\draw[webNo, line width=0.8pt, line cap=round] (0,2.6pt) -- (5.2pt,2.6pt);}%
\def\Y{\usebox{\relYes}}\def\N{\usebox{\relNo}}\def\Yo{\usebox{\relOwn}}%
\newlength{\relQ}\settowidth{\relQ}{TheAgentCompany (Xu et al., 2024b)}%
\newlength{\relK}\setlength{\relK}{\dimexpr(\textwidth-\relQ-26pt)/7\relax}%
\newcolumntype{Q}{>{\raggedright\arraybackslash}p{\relQ}}%
\newcolumntype{K}[1]{>{\centering\arraybackslash}p{#1\relK}}%
\setlength{\tabcolsep}{0pt}%
\newcommand{\thd}[1]{\textcolor{webGraphite}{\scriptsize\bfseries #1}}%
\newcommand{\grp}[1]{\multicolumn{8}{@{\hspace{12pt}}l}{\rule{0pt}{2.75ex}\scriptsize\bfseries\textcolor{webGraphite}{#1}}}%
\newcommand{\hair}{\arrayrulecolor{webHairline}\specialrule{0.4pt}{0pt}{0pt}}%
\newsavebox{\relHead}\newsavebox{\relBody}\newsavebox{\relRow}%
\sbox{\relHead}{\renewcommand{\arraystretch}{1.0}\begin{tabular}{@{\hspace{12pt}}QK{1.05}K{1.0}K{0.9}K{1.12}K{1.18}K{0.95}K{0.8}@{\hspace{14pt}}}
\rule{0pt}{3.0ex}\thd{Benchmark} & \thd{\#Physical} & \thd{Real work} & \thd{Expert} & \thd{Large input} & \thd{Large output} & \thd{Objective} & \thd{Rubric} \\
\rule[-1.55ex]{0pt}{0pt} & \thd{eng.\ tasks} & \thd{tasks} & \thd{curation} & \thd{corpus} & \thd{artifacts} & \thd{function} & \\
\end{tabular}}%
\sbox{\relBody}{\renewcommand{\arraystretch}{1.27}\begin{tabular}{@{\hspace{12pt}}QK{1.05}K{1.0}K{0.9}K{1.12}K{1.18}K{0.95}K{0.8}@{\hspace{14pt}}}
\grp{Software \& ML engineering} \\
SWE-bench~\citep{jimenez2023swebench}         & \N & \Y & \N & \N & \Y & \Y & \N \\ \hair
MLE-bench~\citep{chan2024mlebench}            & \N & \N & \N & \N & \Y & \Y & \N \\ \hair
\grp{Professional work} \\
TheAgentCompany~\citep{xu2024theagentcompany} & \N & \Y & \N & \Y & \Y & \Y & \N \\ \hair
GDPval~\citep{patwardhan2025gdpval}           & \N & \Y & \Y & \Y & \Y & \N & \Y \\ \hair
APEX-Agents~\citep{vidgen2026apexagents}      & \N & \Y & \Y & \Y & \Y & \N & \Y \\ \hair
\grp{Professional physical engineering} \\
EngiBench~\citep{zhou2025engibench}           & 143   & \N & \N & \N & \N & \Y & \Y \\ \hair
EngDesign~\citep{guo2025engdesign}            & 13    & \N & \Y & \N & \N & \Y & \N \\ \hair
Frontier-Eng~\citep{chi2026frontiereng}       & 3     & \N & \N & \N & \Y & \Y & \N \\ \hair
DrafterBench~\citep{li2025drafterbench}       & 1,920 & \N & \N & \N & \Y & \Y & \N \\ \hair
ElecBench~\citep{zhou2024elecbench}           & 288   & \N & \N & \N & \N & \Y & \N \\
\end{tabular}}%
\sbox{\relRow}{\renewcommand{\arraystretch}{1}\begin{tabular}{@{\hspace{12pt}}QK{1.05}K{1.0}K{0.9}K{1.12}K{1.18}K{0.95}K{0.8}@{\hspace{14pt}}}
\rule[-1.1ex]{0pt}{3.2ex}\textbf{\bench{} (ours)} & \textbf{85} & \Yo & \Yo & \Yo & \Yo & \Yo & \Yo \\
\end{tabular}}%
\newlength{\relHh}\setlength{\relHh}{\dimexpr\ht\relHead+\dp\relHead\relax}%
\newlength{\relBh}\setlength{\relBh}{\dimexpr\ht\relBody+\dp\relBody\relax}%
\newlength{\relRh}\setlength{\relRh}{\dimexpr\ht\relRow+\dp\relRow\relax}%
\newlength{\relW}\setlength{\relW}{\wd\relHead}%
\begin{tikzpicture}[x=1pt, y=1pt]
\begin{scope}
  \clip[rounded corners=6pt] (0,0) rectangle (\relW,{-\relHh-\relBh-\relRh-6pt});
  \fill[white] (0,0) rectangle (\relW,{-\relHh-\relBh-\relRh-6pt});
  \fill[webWash] (0,0) rectangle (\relW,-\relHh);
\end{scope}
\draw[webHairline, line width=0.4pt] (0,-\relHh) -- (\relW,-\relHh);
\fill[webTint, rounded corners={0.5\relRh}] (4pt,{-\relHh-\relBh-2pt}) rectangle (\relW-4pt,{-\relHh-\relBh-2pt-\relRh});
\node[inner sep=0pt, anchor=north west] at (0,0) {\usebox{\relHead}};
\node[inner sep=0pt, anchor=north west] at (0,-\relHh) {\usebox{\relBody}};
\node[inner sep=0pt, anchor=north west] at (0,{-\relHh-\relBh-2pt}) {\usebox{\relRow}};
\draw[webHairline, line width=0.5pt, rounded corners=6pt] (0,0) rectangle (\relW,{-\relHh-\relBh-\relRh-6pt});
\end{tikzpicture}%
\arrayrulecolor{pleHairline}
\end{table}

\paragraph{Agentic benchmarks for software and ML engineering.}
The closest agentic benchmarks evaluate digital engineering domains. SWE-bench and its
variants~\citep{jimenez2023swebench,openai2024swebenchverified,yang2024swebenchmm,miserendino2025swelancer}
measure software engineering on real repositories, and MLE-bench~\citep{chan2024mlebench} and
RE-Bench~\citep{wijk2024rebench} measure machine-learning engineering; related suites cover
terminals~\citep{merrill2026terminalbench}, operating systems and the
web~\citep{xie2024osworld,zhou2023webarena,mialon2023gaia}, tool--user
interaction~\citep{yao2024taubench}, security~\citep{zhang2024cybench}, scientific
code~\citep{tian2024scicode} and paper replication~\citep{starace2025paperbench}. These benchmarks
established the sandboxed, tool-calling protocol that we also adopt, following code-as-action
agents~\citep{yao2022react,wang2024codeact,wang2024openhands}. Their artifacts, however, are purely digital:
code, models and their test suites, where success is a unit-test pass or a held-out metric.
Professional physical engineering instead produces designs for physical systems, governed by laws of physics and
legally binding building codes, and documented in production drawings, schedules and calculations that a licensed reviewer signs off on.

\paragraph{Benchmarks of professional work.}
A recent line of work measures economically valuable knowledge work.
TheAgentCompany~\citep{xu2024theagentcompany} simulates a software company.
GDPval~\citep{patwardhan2025gdpval} collects deliverable-producing tasks from experienced
professionals in 44 occupations and grades them by blinded expert pairwise comparison.
APEX~\citep{vidgen2025apex} and APEX-Agents~\citep{vidgen2026apexagents} have banking,
consulting, legal and medical experts author tasks and rubrics. APEX-Agents runs them in
long-horizon, cross-application ``environments'' through the Archipelago
harness~\citep{mercor2026archipelago}, which our environment builds on. HealthBench~\citep{arora2025healthbench} and
PaperBench~\citep{starace2025paperbench} show that large expert rubrics, graded by a validated
LLM judge~\citep{zheng2023judging,kim2023prometheus,gu2024llmjudgesurvey}, scale open-ended
evaluation. \bench{} brings this focus on workplace realism and direct economic value to professional physical engineering: it adopts expert rubrics for its expert-authored tasks, while complementing half of the benchmark with an objective, programmatic measure of engineering design quality and cost efficiency.

\paragraph{Professional physical engineering.}
Few benchmarks touch physical engineering, and only in small slices (\cref{tab:related}):
143 of EngiBench's 1,760 textbook-style problems~\citep{zhou2025engibench}, 13 generic member-sizing
and topology tasks in EngDesign~\citep{guo2025engdesign}, and 3 of Frontier-Eng's 47 simulator-based
optimisation tasks~\citep{chi2026frontiereng}. ElecBench~\citep{zhou2024elecbench} and
DrafterBench~\citep{li2025drafterbench} cover narrow sub-tasks (power-system question answering and
single-sheet drawing revisions), and building services and structures have LLM
systems~\citep{liang2025masse} and reviews~\citep{neubauer2026hvacllm} but no agentic benchmark.
These benchmarks abstract away the multi-document project corpus and production deliverables of
real practice, and they are close to saturation: the best agent beats Frontier-Eng's baseline on
97.9\% of its tasks, and leading models exceed 80\% at EngiBench Level~2. \bench{} instead tests
whether an agent can carry a professional job from raw project documents to a deliverable package
checked against codes, physics and economic optimality (\cref{sec:coauth}). Testing co-authored tasks
against a frontier solver and keeping part of the benchmark private, like unpublished problem
sets~\citep{glazer2024frontiermath}, guard \bench{} against saturation and training-data
contamination~\citep{xu2024contamination,akhtar2026saturation}.

\section{Data Curation}
\label{sec:data}

A task in \bench{} is designed to mirror a real, economically valuable engineering assignment rather than an academic exercise, and consists of four parts. The \emph{instruction} is the brief a professional physical engineer
would receive on an active project. The \emph{environment} is a realistic project file system that holds the multi-format engineering corpus the agent
must work from. The \emph{deliverable specification} names the production work-product files the agent must issue so that real engineering design work gets done. The
\emph{grader} is either a weighted expert rubric or an executable verifier with withheld
reference deliverables. All tasks are packaged in the APEX format~\citep{vidgen2026apexagents} and run in
Archipelago-style environment containers~\citep{mercor2026archipelago}. These expose the environment
through MCP applications (filesystem, documents, spreadsheets, PDFs, mail,
calendar, chat) and a code-execution sandbox.

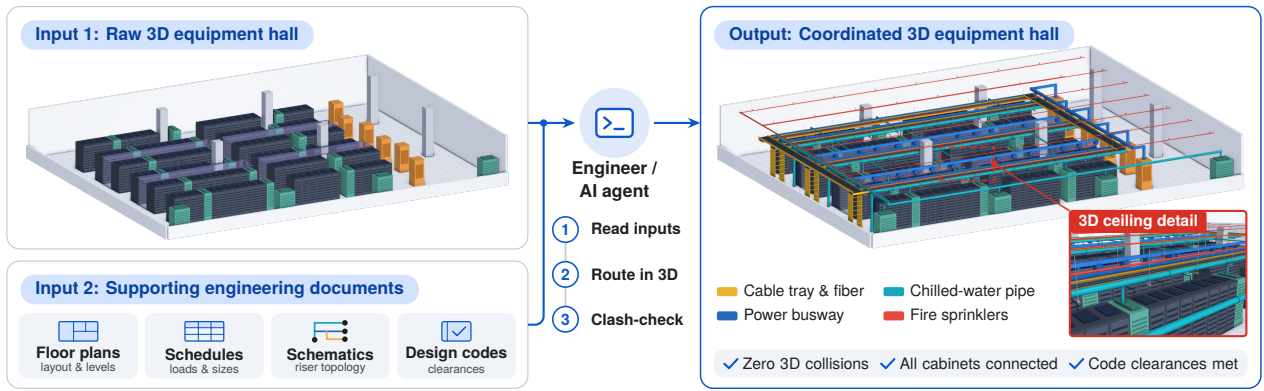
\begin{figure*}[t]
\centering
\definecolor{webInk}{HTML}{1F1F1F}\definecolor{webGraphite}{HTML}{444746}%
\definecolor{webHairline}{HTML}{C4C7C5}\definecolor{webWash}{HTML}{F0F4F9}%
\definecolor{webBlue}{HTML}{0B57D0}\definecolor{webPill}{HTML}{D3E3FD}\definecolor{webPillText}{HTML}{0842A0}%
\definecolor{webRed}{HTML}{D93025}\colorlet{webTint}{webBlue!13}%
\definecolor{f3tray}{HTML}{E9B22F}\definecolor{f3bus}{HTML}{2868BA}%
\definecolor{f3chw}{HTML}{17A6B8}\definecolor{f3spr}{HTML}{E4473C}%
\sffamily
\resizebox{\textwidth}{!}{%
\begin{tikzpicture}[
  card/.style={draw=webHairline, line width=0.5pt, fill=white, rounded corners=6pt},
  outcard/.style={draw=webBlue, line width=0.7pt, fill=white, rounded corners=6pt},
  pill/.style={fill=webPill, rounded corners=6.5pt, font=\footnotesize\bfseries, text=webPillText,
               inner xsep=7pt, inner ysep=3pt, anchor=north west},
  doc/.style={fill=webWash, rounded corners=4pt},
  doclbl/.style={font=\scriptsize\bfseries, text=webInk, align=center},
  docsub/.style={font=\tiny, text=webGraphite, align=center},
  icon/.style={draw=webBlue, line width=0.6pt, line join=round, line cap=round},
  stepdot/.style={circle, draw=webBlue, line width=0.75pt, fill=white, minimum size=0.42cm, inner sep=0pt,
               font=\scriptsize\bfseries, text=webBlue},
  steplbl/.style={font=\scriptsize\bfseries, text=webInk, anchor=west},
  flow/.style={-{Stealth[length=2.1mm,width=1.7mm]}, line width=1pt, draw=webBlue, rounded corners=4pt},
  key/.style={font=\scriptsize, text=webInk, anchor=west, inner sep=1pt},
  checklbl/.style={font=\scriptsize, text=webInk, inner sep=1pt},
]
\def\p{figures/task_3d}
\newcommand{\fthreetick}{\tikz[baseline=0pt]\draw[webBlue, line width=1pt, line cap=round, line join=round]
  (0,2.6pt) -- (2.2pt,0.5pt) -- (6.2pt,4.8pt);\hspace{0.28em}}

\draw[card] (0,2.30) rectangle (8.60,6.30);
\node[pill] at (0.22,6.08) {Input 1: Raw 3D equipment hall};
\node[inner sep=0pt] at (4.30,3.98) {\includegraphics[width=8.05cm]{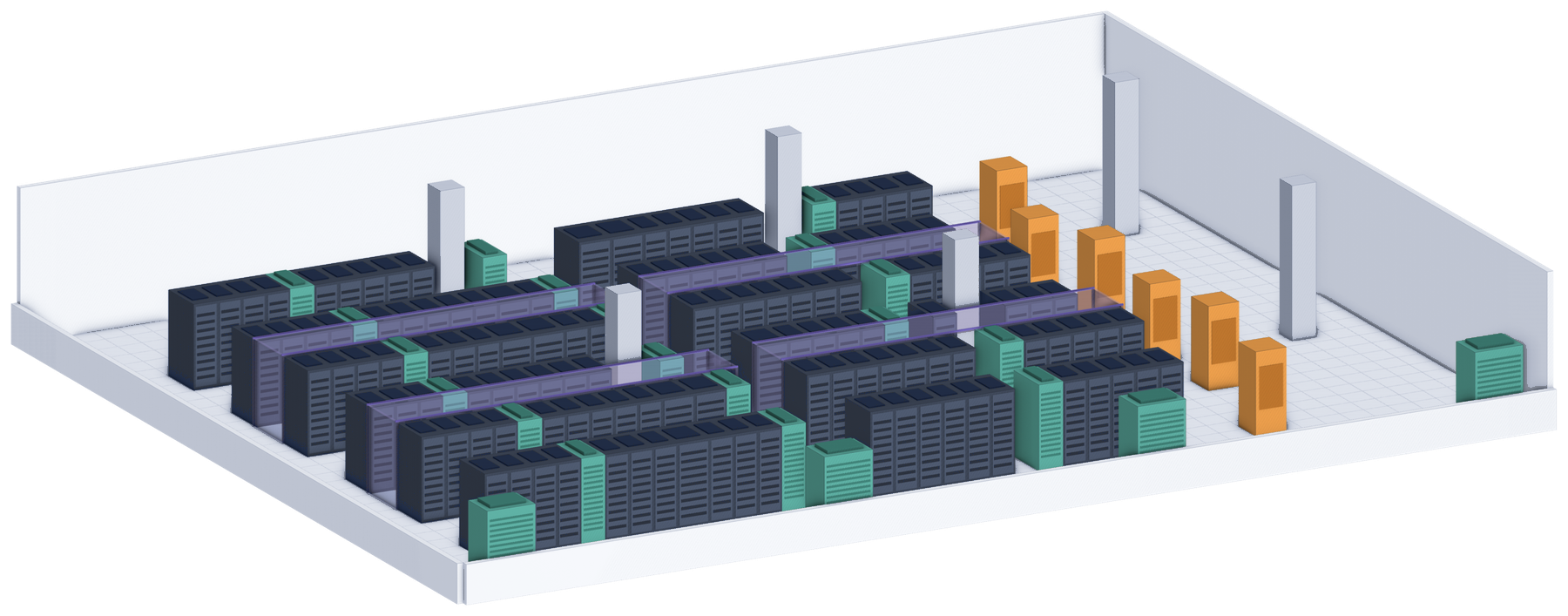}};

\draw[card] (0,0) rectangle (8.60,2.10);
\node[pill] at (0.22,1.88) {Input 2: Supporting engineering documents};
\foreach \x in {0.20,2.28,4.36,6.44} { \fill[doc] (\x,0.16) rectangle ++(1.96,1.08); }
\begin{scope}[shift={(0.86,0.78)}]   
  \draw[icon, fill=webPill, rounded corners=0.6pt] (0,0) rectangle (0.64,0.32);
  \draw[icon, line width=0.4pt] (0.24,0) -- (0.24,0.32) (0.24,0.16) -- (0.64,0.16) (0.44,0.16) -- (0.44,0.32);
\end{scope}
\node[doclbl] at (1.18,0.56) {Floor plans};
\node[docsub] at (1.18,0.33) {layout \& levels};
\begin{scope}[shift={(2.94,0.78)}]   
  \fill[webPill] (0,0.22) rectangle (0.64,0.32);
  \draw[icon, rounded corners=0.6pt] (0,0) rectangle (0.64,0.32);
  \draw[icon, line width=0.4pt] (0,0.11) -- (0.64,0.11) (0,0.22) -- (0.64,0.22) (0.22,0) -- (0.22,0.32) (0.43,0) -- (0.43,0.32);
\end{scope}
\node[doclbl] at (3.26,0.56) {Schedules};
\node[docsub] at (3.26,0.33) {loads \& sizes};
\begin{scope}[shift={(5.02,0.78)}]   
  \draw[f3chw, line width=0.8pt, line cap=round] (0.06,0.02) -- (0.06,0.30) -- (0.56,0.30);
  \draw[f3bus, line width=0.8pt, line cap=round] (0.06,0.16) -- (0.56,0.16);
  \draw[f3tray, line width=0.8pt, line cap=round] (0.26,0.16) -- (0.26,0.03) -- (0.56,0.03);
  \fill[webInk] (0.06,0.02) circle (1.3pt) (0.56,0.30) circle (1.3pt) (0.56,0.16) circle (1.3pt) (0.56,0.03) circle (1.3pt);
\end{scope}
\node[doclbl] at (5.34,0.56) {Schematics};
\node[docsub] at (5.34,0.33) {riser topology};
\begin{scope}[shift={(7.14,0.78)}]   
  \draw[icon, fill=webPill, rounded corners=1pt] (0.04,0) rectangle (0.52,0.32);
  \draw[icon, line width=0.45pt] (0.13,0) -- (0.13,0.32);
  \draw[icon, line width=0.7pt] (0.22,0.16) -- (0.29,0.09) -- (0.43,0.23);
\end{scope}
\node[doclbl] at (7.42,0.56) {Design codes};
\node[docsub] at (7.42,0.33) {clearances};

\fill[webTint] (10.03,4.38) circle (0.58);
\begin{scope}[shift={(10.03,4.38)}, scale=2.0]
  \draw[icon, line width=1.1pt, fill=white, rounded corners=2pt] (-0.15,-0.115) rectangle (0.15,0.115);
  \draw[icon, line width=1.1pt] (-0.085,0.045) -- (-0.03,0) -- (-0.085,-0.045);
  \draw[icon, line width=1.1pt] (0.0,-0.05) -- (0.075,-0.05);
\end{scope}
\node[font=\footnotesize\bfseries, text=webInk, align=center] at (10.03,3.42) {Engineer /\\[-0.5pt] AI agent};
\draw[webHairline, line width=0.75pt] (9.22,2.62) -- (9.22,1.14);
\node[stepdot] at (9.22,2.62) {1}; \node[steplbl] at (9.52,2.62) {Read inputs};
\node[stepdot] at (9.22,1.88) {2}; \node[steplbl] at (9.52,1.88) {Route in 3D};
\node[stepdot] at (9.22,1.14) {3}; \node[steplbl] at (9.52,1.14) {Clash-check};

\draw[flow] (8.60,4.38) -- (9.38,4.38);
\draw[line width=1pt, draw=webBlue, rounded corners=4pt] (8.60,1.05) -- (8.86,1.05) -- (8.86,4.38);
\fill[webBlue] (8.86,4.38) circle (1.7pt);
\draw[flow] (10.68,4.38) -- (11.45,4.38);

\draw[outcard] (11.45,0) rectangle (20.70,6.30);
\node[pill] at (11.67,6.08) {Output: Coordinated 3D equipment hall};
\node[inner sep=0pt] at (16.08,3.98) {\includegraphics[width=8.80cm]{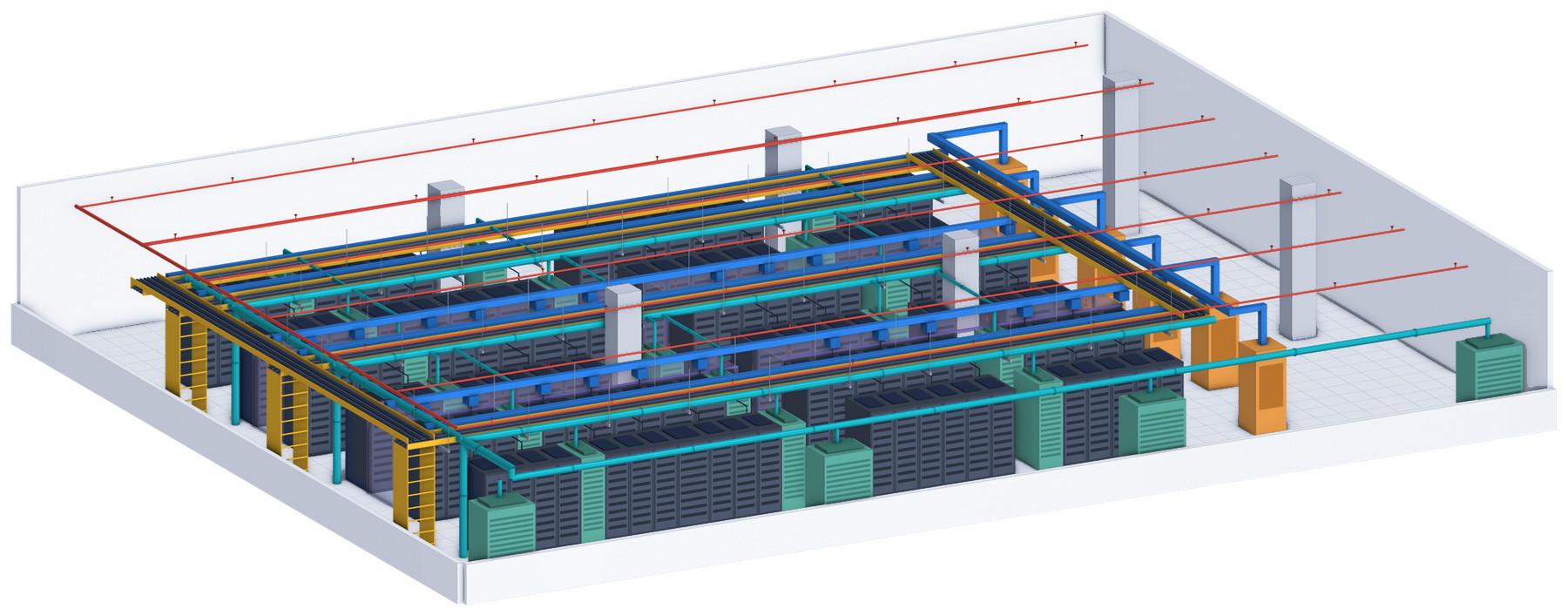}};

\begin{scope}[shift={(11.72,1.16)}]
  \fill[f3tray, rounded corners=1pt] (0,0.36) rectangle ++(0.34,0.15); \node[key] at (0.40,0.435) {Cable tray \& fiber};
  \fill[f3bus,  rounded corners=1pt] (0,-0.04) rectangle ++(0.34,0.15); \node[key] at (0.40,0.035) {Power busway};
  \fill[f3chw,  rounded corners=1pt] (2.75,0.36) rectangle ++(0.34,0.15); \node[key] at (3.15,0.435) {Chilled-water pipe};
  \fill[f3spr,  rounded corners=1pt] (2.75,-0.04) rectangle ++(0.34,0.15); \node[key] at (3.15,0.035) {Fire sprinklers};
\end{scope}

\draw[webRed, densely dashed, line width=0.75pt] (15.40,3.43) rectangle (16.30,3.98);
\node[inner sep=0pt, anchor=south east] (det) at (20.46,0.86) {\includegraphics[width=2.90cm]{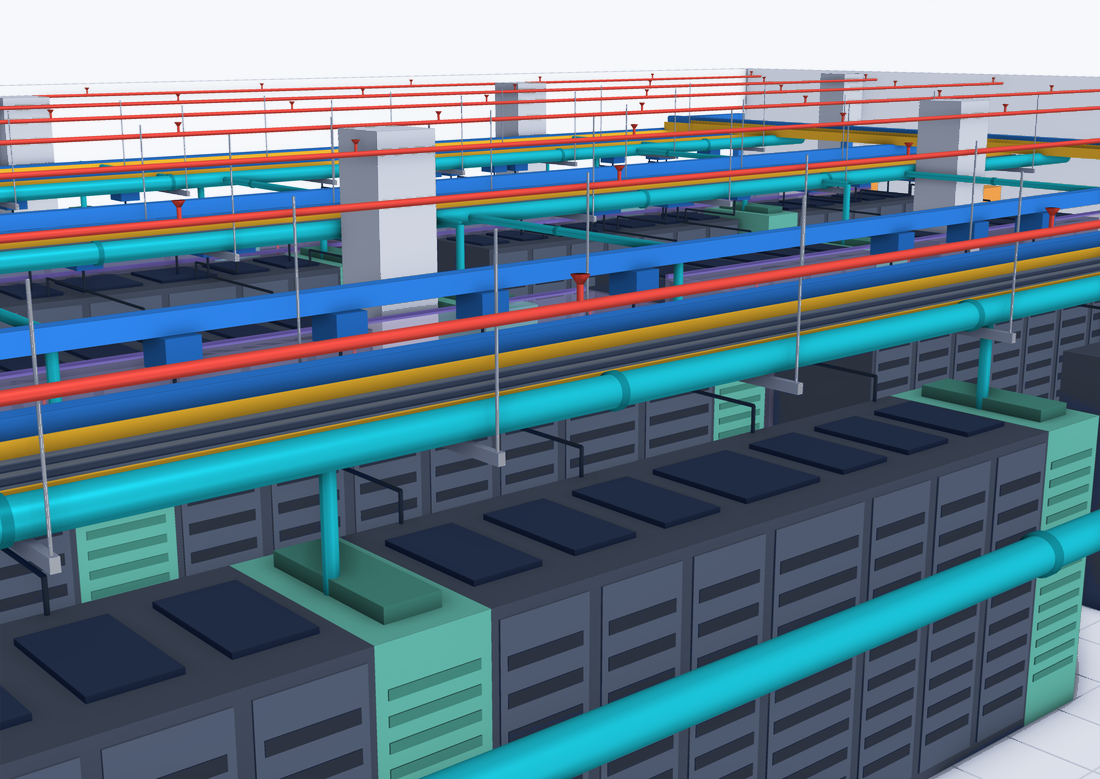}};
\draw[webRed, line width=1pt, rounded corners=1.5pt] (det.south west) rectangle (det.north east);
\draw[webRed, line width=0.7pt] (16.30,3.70) -- (det.north west);
\fill[webRed] (16.30,3.70) circle (1.6pt);
\node[fill=webRed, text=white, font=\scriptsize\bfseries, rounded corners=1.5pt, inner xsep=4pt, inner ysep=1.6pt,
      anchor=north west] at (det.north west) {3D ceiling detail};

\fill[webWash, rounded corners=4pt] (11.67,0.18) rectangle (20.48,0.64);
\node[checklbl, text width=8.50cm, align=left] at (16.075,0.41)
  {\fthreetick Zero 3D collisions\hspace{0.6em plus 1fill}\fthreetick All cabinets connected\hspace{0.6em plus 1fill}\fthreetick Code clearances met};

\end{tikzpicture}}
\caption{An example \bench{} 3D task: multi-system MEP coordination of an equipment hall (\emph{MEP coordination, hall A} with 8~input files).
\textbf{Input 1:} A raw 3D equipment hall model with floor equipment in place but no overhead distribution.
\textbf{Input 2:} Supporting engineering documents (floor plans, schedules, schematics and design codes).
\textbf{Output:} A complete, collision-free 3D model routing overhead cable trays, power busways, chilled-water piping and fire sprinklers to connect every cabinet while satisfying code clearances and minimising route length.}
\label{fig:task_example}
\end{figure*}

\paragraph{Detailed example.}
\Cref{fig:task_example} shows a concrete task from \bench{} (\emph{MEP coordination, hall A} with 8~input files): an MEP coordinator is handed the drawing set, schedules, schematics, design codes and raw 3D model of an
equipment hall with 128 equipment cabinets and asked to deliver a fully coordinated 3D model. The engineer
must read positions off the plans, sizes off the schedules and elevations off the section, and
reconcile conflicts across the sheets. They must then route the cable tray, power busway, chilled-water mains and fire sprinklers
through the ceiling space so that nothing clashes with the cabinets, columns, containment or
sprinkler heads, every cabinet row and cooling unit is served, the sprinkler and electrical code
clearances hold, and the routes are as short and economical as practical. The deliverable is a production-ready 3D model that a
licensed reviewer and our executable grader can clash-check and measure against the governing documents, physical clearances and routing-length objective.

The tasks in \bench{} come from two curation approaches. \emph{Expert-Authored} tasks are written end to end by
practising engineers (\cref{sec:expert}). \emph{Co-Authored} tasks start from briefs written by
practising engineers, are built by our agent pipeline, and are then validated and have their difficulty intensified
by practising engineers (\cref{sec:coauth}).

\subsection{Expert-Authored Tasks}
\label{sec:expert}

Our first route relies entirely on human expertise to capture authentic workplace realism and billable engineering workflows. We collaborated with Mercor, which
maintains a large pool of vetted domain professionals and has used it to build expert
benchmarks of professional work~\citep{vidgen2025apex,vidgen2026apexagents}. For \bench{},
Mercor recruited practising civil, electrical and mechanical engineers. Each expert built a task in three stages.

\paragraph{Task authoring.}
Experts first write the task as a realistic work request from a project colleague or client. The request describes the
engineering situation with its constraints and specifies the exact deliverables needed to advance the project. For example:
\begin{quote}\small
``Garrett wants the Phase~3 estimate out to the surety broker and I don't want it going
anywhere until Section~2 has been checked against what's actually on the plans. [\ldots]
Give me the corrected Section~2 subtotal, a line-by-line table of every quantity or
unit-price change with the sheet it comes off, and the revised rounded bond figure. [\ldots]
Put it in a memo to Callaway for my signature, dated 10/31/2026, saved as
\texttt{Correspondence/Client/Memo\_to\_Callaway\_2026-10-31.docx}.''
\end{quote}
Answering this request requires cross-referencing a cost estimate (XLSX), three drainage sheets
of a nine-sheet plan set (PDF) and a hydraulic-grade-line report. The solver must then apply
the stated pricing rules and write a signed memo. The expert-authored tasks cover
plan-review responses, submittal reviews, capacity and energization decisions, geotechnical
and foundation checks, tank shell-thickness review, close-out letters and cost positions.

\paragraph{Environment building.}
The documents a request depends on live in its \emph{environment}: a realistic project file system of
the kind an engineer inherits on joining a job. Experts build the environment around the task. It
contains drawing sets, calculation reports, equipment and
motor schedules, test reports, cost estimates, meeting minutes, e-mail threads and
superseded revisions, organised in folders such as \texttt{Design-Drafting/},
\texttt{Cost Estimates/}, \texttt{Reports-Studies/} and
\texttt{Correspondence/}. Each task comes with 63 environment files on average (range 50--112) in six native formats,
mostly PDF and XLSX (\cref{tab:stats}). The 44 tasks share 12 environments, so a shared environment counts once for
each task that uses it.

\paragraph{Grading.}
Each task is graded by a weighted rubric of binary \emph{verifiers}. Every verifier takes as
input the deliverable file it reads, a single checkable criterion (e.g.\ ``the memorandum states that the
corrected Section~2 subtotal is \$1,303,717.70''), a severity and a numerical weight.
Load-bearing numbers carry high weights (10 in the example above) and presentation items
carry low ones (1). The 44 expert-authored tasks carry 941 rubric items, 21.4 per task on average.
A task's score is the weight-normalised fraction of verifiers passed, as judged by an LLM
that reads the deliverable file itself. Each task went through Mercor's quality control before delivery.

\subsection{Co-Authored Tasks}
\label{sec:coauth}

\begin{figure*}[t]
\centering
\definecolor{webInk}{HTML}{1F1F1F}%
\definecolor{webGraphite}{HTML}{444746}%
\definecolor{webHairline}{HTML}{C4C7C5}%
\definecolor{webWash}{HTML}{F0F4F9}%
\definecolor{webBlue}{HTML}{0B57D0}%
\definecolor{webRed}{HTML}{D93025}%
\colorlet{webTint}{webBlue!13}
\colorlet{webTrack}{webGraphite!18}
\def\stH{4.30}
\newcommand{\stage}[4]{
  \draw[sheet] (#1,0) rectangle (#2,\stH);
  \node[balloon] at (#1+0.34,\stH-0.34) {#3};
  \node[stagetitle] at (#1+0.64,\stH-0.34) {#4};
  \draw[webBlue, line width=0.5pt] (#1,\stH-0.68) -- (#2,\stH-0.68);}
\newcommand{\actor}[5]{
  \node[badge] at (#1+0.42,\stH-1.12) {};
  \begin{scope}[shift={(#1+0.42,\stH-1.12)}]#3\end{scope}
  \node[actor] at (#1+0.84,\stH-0.99) {#4};
  \node[sub] at (#1+0.84,\stH-1.13) {#5};
  \draw[hair] (#1,\stH-1.56) -- (#2,\stH-1.56);}
\newcommand{\itemrow}[4]{
  \begin{scope}[shift={(#1+0.42,#2)}]#3\end{scope}
  \node[item] at (#1+0.84,#2+0.17) {#4};}
\newcommand{\itemwhat}[3]{\node[sub] at (#1+0.84,#2+0.02) {#3};}
\newcommand{\engineer}{
  \fill[webBlue] (-0.17,-0.20) .. controls (-0.16,-0.06) and (0.16,-0.06) .. (0.17,-0.20) -- cycle;
  \draw[icon, fill=white] (0,0.01) circle (0.085);
  \fill[webBlue] (-0.10,0.05) arc (180:0:0.10 and 0.095) -- cycle;
  \draw[webBlue, line width=0.8pt, line cap=round] (-0.135,0.05) -- (0.135,0.05);}
\newcommand{\terminal}{
  \draw[icon, fill=white, rounded corners=1.2pt] (-0.15,-0.115) rectangle (0.15,0.115);
  \draw[icon] (-0.085,0.045) -- (-0.03,0) -- (-0.085,-0.045);
  \draw[icon] (0.0,-0.05) -- (0.075,-0.05);}
\newcommand{\sandboxed}{
  \draw[webBlue, line width=0.5pt, dash pattern=on 1.1pt off 0.9pt, rounded corners=1.5pt] (-0.19,-0.155) rectangle (0.19,0.155);
  \draw[icon, fill=white, rounded corners=1pt] (-0.12,-0.09) rectangle (0.12,0.09);
  \draw[icon, line width=0.55pt] (-0.065,0.035) -- (-0.022,0) -- (-0.065,-0.035);
  \draw[icon, line width=0.55pt] (0.0,-0.04) -- (0.06,-0.04);}
\newcommand{\docicon}{\draw[icon, fill=white, rounded corners=1pt] (-0.13,-0.17) rectangle (0.13,0.17);
  \draw[icon, line width=0.5pt] (-0.07,0.08) -- (0.07,0.08) (-0.07,0.0) -- (0.07,0.0) (-0.07,-0.08) -- (0.02,-0.08);}
\newcommand{\scopeicon}{\draw[icon, fill=white, rounded corners=1pt] (-0.15,-0.15) rectangle (0.15,0.15);
  \draw[icon] (-0.075,0.0) -- (-0.02,-0.06) -- (0.08,0.065);}
\newcommand{\foldericon}{\draw[icon, fill=white, rounded corners=0.8pt]
  (-0.16,-0.13) -- (-0.16,0.13) -- (-0.05,0.13) -- (-0.01,0.08) -- (0.16,0.08) -- (0.16,-0.13) -- cycle;
  \draw[icon, line width=0.5pt] (-0.16,0.03) -- (0.16,0.03);}
\newcommand{\verifyicon}{\draw[icon, fill=white] (0,0.17) -- (0.15,0.11) -- (0.15,0.0) .. controls (0.15,-0.09) and (0.07,-0.15) .. (0,-0.18)
  .. controls (-0.07,-0.15) and (-0.15,-0.09) .. (-0.15,0.0) -- (-0.15,0.11) -- cycle;
  \draw[icon] (-0.07,0.0) -- (-0.02,-0.055) -- (0.075,0.055);}
\newcommand{\signtick}{\tikz[baseline=0pt]\draw[webBlue, line width=0.9pt, line cap=round, line join=round]
  (0,2.5pt) -- (2.1pt,0.4pt) -- (6.0pt,4.6pt);}

\resizebox{\textwidth}{!}{%
\begin{tikzpicture}[x=1cm, y=1cm, font=\sffamily,
  sheet/.style={draw=webBlue, line width=0.6pt, fill=white, rounded corners=6pt},
  balloon/.style={circle, draw=webBlue, line width=0.75pt, fill=white, inner sep=0pt, minimum size=0.40cm,
    font=\sffamily\fontsize{7}{8}\selectfont\bfseries, text=webBlue},
  stagetitle/.style={anchor=west, inner sep=0pt, font=\sffamily\fontsize{8.6}{10}\selectfont\bfseries, text=webInk},
  actor/.style={anchor=west, inner sep=0pt, font=\sffamily\fontsize{7.6}{9}\selectfont\bfseries, text=webInk},
  sub/.style={anchor=north west, inner sep=0pt, font=\sffamily\fontsize{6.6}{8}\selectfont, text=webGraphite, align=left},
  item/.style={anchor=west, inner sep=0pt, font=\sffamily\fontsize{7.3}{8.6}\selectfont\bfseries, text=webInk},
  hair/.style={draw=webHairline, line width=0.45pt},
  flow/.style={-{Stealth[length=2.3mm,width=1.9mm]}, line width=1.15pt, draw=webBlue},
  msg/.style={font=\sffamily\fontsize{6.8}{7.8}\selectfont\bfseries, text=webGraphite, inner sep=0pt, align=center},
  icon/.style={draw=webBlue, line width=0.65pt, line cap=round, line join=round},
  badge/.style={circle, fill=webTint, inner sep=0pt, minimum size=0.56cm},
]
\fill[webWash, rounded corners=9pt] (4.35,-1.02) rectangle (14.25,\stH+0.72);
\node[anchor=west, rounded corners=5.5pt, fill=webTint, text=webBlue, inner xsep=6.5pt, inner ysep=2.6pt,
  font=\sffamily\fontsize{7.2}{8.4}\selectfont\bfseries] at (4.6,\stH+0.36) {Adversarial Creator--Solver Co-Curation Loop};

\stage{0}{3.6}{1}{Expert Seed Prompt}
\actor{0}{3.6}{\engineer}{Practising Engineer}{Domain task nomination}
\itemrow{0}{2.0}{\docicon}{Problem Brief}
\itemwhat{0}{2.0}{Real design scenario\\\& governing codes}
\draw[hair] (0,1.37) -- (3.6,1.37);
\itemrow{0}{0.72}{\scopeicon}{Target Scope}
\itemwhat{0}{0.72}{Required engineering\\deliverable specs}

\draw[flow] (3.6,2.15) -- (4.6,2.15);
\node[msg, anchor=south] at (4.1,2.27) {Seed};

\stage{4.6}{8.55}{2}{Task Creator Agent}
\actor{4.6}{8.55}{\terminal}{Creator Agent}{Deep research \& synthesis}
\itemrow{4.6}{2.17}{\docicon}{Task Prompt}
\itemwhat{4.6}{2.17}{Actionable design specs}
\draw[hair] (4.6,1.70) -- (8.55,1.70);
\itemrow{4.6}{1.25}{\foldericon}{Project Environment}
\itemwhat{4.6}{1.25}{Multi-format docs \& CAD}
\draw[hair] (4.6,0.82) -- (8.55,0.82);
\itemrow{4.6}{0.40}{\verifyicon}{Verifier \& Reference}
\itemwhat{4.6}{0.40}{Physics, code \& cost checks}

\draw[flow] (8.55,2.15) -- (9.65,2.15);
\node[msg, anchor=south] at (9.1,2.27) {Task\\[-1pt]package};

\stage{9.65}{14.0}{3}{Task Solver Agent}
\actor{9.65}{14.0}{\sandboxed}{Solver Agent ($N{=}3$)}{Sandbox w/o verifier or reference}
\node[item] at (9.87,2.43) {Difficulty Gate};
\foreach \r/\y/\s in {1/2.02/0.44, 2/1.70/0.36, 3/1.38/0.40} {
  \node[anchor=west, inner sep=0pt, font=\sffamily\fontsize{6.4}{7.4}\selectfont, text=webGraphite] at (9.87,\y) {Run \r};
  \fill[webTrack, rounded corners=0.6pt] (10.55,\y-0.055) rectangle (13.75,\y+0.055);
  \fill[webBlue, rounded corners=0.6pt] (10.55,\y-0.055) rectangle ({10.55+\s*3.2},\y+0.055);
}
\draw[webRed, line width=0.9pt, dash pattern=on 2pt off 1.4pt] (12.73,1.22) -- (12.73,2.20);
\node[anchor=south, inner sep=0pt, font=\sffamily\fontsize{6.4}{7.4}\selectfont\bfseries, text=webRed] at (12.73,2.27) {Target $\tau$};
\draw[hair] (9.65,1.06) -- (14.0,1.06);
\node[anchor=west, inner sep=0pt, font=\sffamily\fontsize{6.6}{8}\selectfont, text=webInk] at (9.87,0.74)
  {\textcolor{webBlue}{\textbf{Pass} ($\bar{s} < \tau$)}: Advance to expert review};
\node[anchor=west, inner sep=0pt, font=\sffamily\fontsize{6.6}{8}\selectfont, text=webInk] at (9.87,0.36)
  {\textcolor{webRed}{\textbf{Fail} ($\bar{s} \ge \tau$)}: Return to Creator loop};

\draw[-{Stealth[length=2.3mm,width=1.9mm]}, line width=1.15pt, draw=webRed, rounded corners=5pt]
  (11.83,0) -- (11.83,-0.55) -- (6.58,-0.55) -- (6.58,0);
\node[fill=white, draw=webRed, line width=0.6pt, rounded corners=1.5pt, inner xsep=5pt, inner ysep=2.6pt,
  text=webRed, font=\sffamily\fontsize{7}{8}\selectfont\bfseries] at (9.205,-0.55) {Inspect Solver Logs \& Intensify Difficulty};

\draw[flow] (14.0,2.15) -- (15.0,2.15);
\node[msg, anchor=south] at (14.5,2.27) {Pass};

\stage{15.0}{18.6}{4}{Expert Review}
\actor{15.0}{18.6}{\engineer}{Licensed Engineer}{Independent verification}
\node[item] at (15.22,2.43) {Expert Sign-Off};
\foreach \y/\t in {2.05/{\textbf{Validate} realism \& reference}, 1.73/{\textbf{Verify} grader \& prompt}, 1.41/{\textbf{Intensify} difficulty}} {
  \node[anchor=west, inner sep=0pt, font=\sffamily\fontsize{6.6}{8}\selectfont, text=webInk] at (15.22,\y) {\signtick\hspace{3.5pt}\t};
}
\fill[webBlue, rounded corners=4pt] (15.2,0.2) rectangle (18.4,1.06);
\node[inner sep=0pt, font=\sffamily\fontsize{7.4}{8.6}\selectfont\bfseries, text=white] at (16.8,0.78) {Released \bench{} Task};
\node[inner sep=0pt, font=\sffamily\fontsize{5.9}{7}\selectfont, text=white] at (16.8,0.45) {Environment, Grader \& Reference};
\end{tikzpicture}%
}
\caption{Overview of the co-authored curation pipeline.
\textbf{(1)~Expert Seed Prompt:} A practising engineer defines the engineering problem brief and target deliverables via our task nomination interface (\cref{fig:nomination_ui}).
\textbf{(2)~Task Creator Agent:} An agent researches real engineering sources to construct the task prompt, multi-format project environment, and executable verifier with a reference solution.
\textbf{(3)~Task Solver Agent:} A sandboxed Solver agent attempts the task across $N{=}3$ runs without verifier or reference access; if the mean score exceeds the difficulty target, the Creator inspects the solver logs to intensify task difficulty.
\textbf{(4)~Expert Review:} Tasks meeting the difficulty target undergo final validation and difficulty intensification by licensed engineers before entering \bench{}.}
\label{fig:pipeline}
\end{figure*}
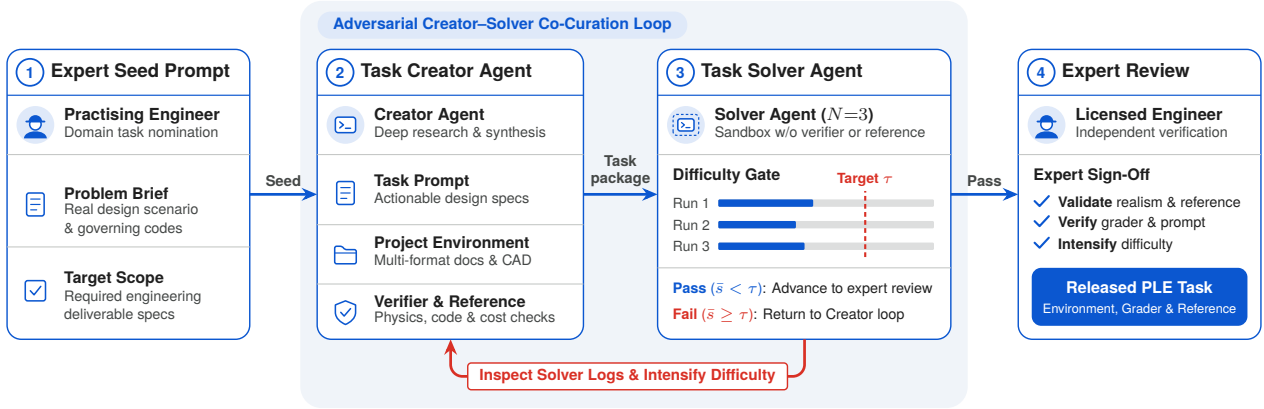

Expert authoring yields realistic tasks, but it scales with expert hours. Our second route
\emph{co-authors} each task between practising engineers and an agent pipeline. Engineers supply
the task brief at the start and validate the finished task at the end; in between, frontier LLM
agents operate under a written method and are checked by programs and independent agents at every
stage. A \emph{Task Creator} agent builds each task
in the same three stages as the experts: task authoring, environment building and grading
(\cref{fig:pipeline}). A \emph{Task Solver} agent then calibrates its difficulty, and an
independent audit checks its ground truth. Mercor's experts finally validate and intensify the difficulty of the
output. Both agents use Claude~Opus~5: the Task Creator runs in Claude Code, and the
Task Solver runs in Hermes Agent~\citep{nous2025hermesagent}, one of the harnesses we use for the baselines (\cref{app:config}).

\paragraph{Task authoring.}
Each co-authored task starts from an \emph{expert seed prompt} submitted by a practising
engineer through our web-based task nomination interface (\cref{fig:nomination_ui}), which captures the engineering situation, workspace inputs, optimisation goal, physical/code constraints, verification method, and expected deliverable format. Based on this seed prompt, the generated task instruction states the engineering concerns and names the governing
documents, but leaves the solver to find the applicable requirements in the environment. Four ground
rules govern creation. (i)~\emph{Ground in reality}: every task is built on real, licensable
artifacts (federal criteria, published standards, public drawing sets, vendor data, site
surveys). (ii)~\emph{Authentic difficulty}: difficulty must come from the inherent complexity of
the artifacts, never from vague instructions or hidden grading traps. (iii)~\emph{Engineered
tasks, not retrieved answers}: the agent is given ingredients such as geometry, properties and
limits, and it must model, integrate or optimise its way to an answer that cannot be looked up.
(iv)~\emph{Verifiable deliverables}: all deliverables can be verified with deterministic
simulators or objective rubrics, against a reference deliverable whose ground truth is itself
audited.

\paragraph{Environment building: deep research.}
The environment is built by a deep-research stage. It first maps the practising engineer's workflow
to the document classes consulted at each step, including a public example of the
deliverable's own industry format. It then harvests a corpus in \emph{native} formats. Each
source's licence is confirmed from the publisher's own page, and an unresolved licence excludes
the source. An accepted environment must have at least three distinct formats, distractors such as
superseded revisions and unrelated sheets, a realistic directory tree
(\texttt{criteria/}, \texttt{specs/}, \texttt{drawings/}, \texttt{vendor/}, \ldots), and
information spread across documents so that the solver must cross-reference them.

\paragraph{Verification and grading.}
Deliverables (CSV or
XLSX schedules, DXF drawings, JSON design records and calculation books) are checked
against independent \emph{constraints} per task, drawn from distinct sources or
distinct physical or regulatory concerns (target 20). The \emph{constraints} mix code limits, physical
feasibility, interfaces between documents, operational and procurement limits, and deliverable
format. Each constraint is tagged \emph{hard} (safety and feasibility) or \emph{soft}
(documentation and conventions), traced to its source row, and checked by the grader. The
solver must also explicitly list the requirements it applied, each with clause and value from
the associated grounding document. The grader checks the listed requirements against the references, so successful navigation of the corpus becomes a metric in the final grading.

Rubric-only scoring, as used for the expert-authored tasks and by Mercor's rubric graders, awards
the weighted fraction of binary criteria met. For design work this has a blind spot. When
both designs fulfil the same set of rubrics, it cannot distinguish a better design from a
merely feasible one. Our co-authored tasks are instead scored on both constraint
compliance and an objective score as
\begin{equation}
\label{eq:score}
s = 0.4\,p + 0.6\,G\,O'
\end{equation}
where $p \in [0,1]$ is the fraction of all constraints passed (as in rubric scoring),
$G \in \{0,1\}$ is the hard-constraint (safety, feasibility) gate, and $O' \in [0,1]$ is the
\emph{objective score}. $G$ is necessary because without it, a design can hack the \emph{objective score} by delivering something unsafe or infeasible, which is detrimental in a real-world task. A task's score is the mean of $s$ over the items of its deliverable.
$O'$ measures how good the design is, and the larger
$0.6$ weight prioritises design quality and hard-constraint satisfaction.
For a task that asks for the cheapest compliant design, the raw objective is
$O = c_{\text{bound}}/c$, where $c$ is the cost of the agent's design
and $c_{\text{bound}}$ is a certified lower bound on the cost of any compliant design.
$O$ lies between 0 and 1 and reaches 1 only when the design is as cheap as the bound. We normalise it against the task's
naive baseline,
\begin{equation}
\label{eq:oprime}
O' = \operatorname{clip}\!\left(\frac{O - O_{\text{naive}}}{1 - O_{\text{naive}}},\, 0,\, 1\right),
\end{equation}
where $O_{\text{naive}}$ is the objective of the un-optimised but compliant design from the task's
calibration ($O_{\text{naive}} = 0$ where a task does not come with a naive design). The intuition of this objective function is that a naive design that fulfils all constraints earns no objective credit, and $O'$ measures the quality lift of a good design from a naive one.

\begin{table}[!t]
\centering
\definecolor{webInk}{HTML}{1F1F1F}%
\definecolor{webGraphite}{HTML}{444746}%
\definecolor{webHairline}{HTML}{C4C7C5}%
\definecolor{webWash}{HTML}{F0F4F9}%
\definecolor{webBlue}{HTML}{0B57D0}%
\definecolor{webBlueLight}{HTML}{7CACF8}%
\newcommand{\swatch}[1]{\textcolor{#1}{\rule[0.05ex]{0.62em}{0.62em}}\hspace{0.32em}}%
\newcommand{\thd}[1]{\textcolor{webGraphite}{\scriptsize\bfseries #1}}%
\newcommand{\hair}{\arrayrulecolor{webHairline}\specialrule{0.4pt}{0pt}{0pt}}%
\newsavebox{\statHead}\newsavebox{\statBody}%
\newlength{\statM}\settowidth{\statM}{\footnotesize\sffamily Checks per task, mean}%
\newlength{\statK}\setlength{\statK}{\dimexpr(0.485\linewidth-\statM-20pt)/3\relax}%
\newcolumntype{M}{>{\raggedright\arraybackslash}p{\statM}}%
\newcolumntype{S}[1]{>{\centering\arraybackslash}p{#1\statK}}%
\newcolumntype{T}{@{\hspace{10pt}}MS{0.86}S{1.28}S{0.86}@{\hspace{10pt}}}%
\newcommand{\grp}[1]{\multicolumn{4}{@{\hspace{10pt}}l}{\rule{0pt}{3.0ex}\scriptsize\bfseries\textcolor{webGraphite}{#1}}}%
\setlength{\tabcolsep}{0pt}%
\sbox{\statHead}{\footnotesize\sffamily\color{webInk}\renewcommand{\arraystretch}{1}\begin{tabular}{T}
\rule[-1.2ex]{0pt}{3.6ex}\thd{Metric} & \thd{\swatch{webBlue}Expert} & \thd{\swatch{webBlueLight}Co-authored} & \thd{All} \\
\end{tabular}}%
\sbox{\statBody}{\footnotesize\sffamily\color{webInk}\renewcommand{\arraystretch}{1.38}\begin{tabular}{T}
Tasks                 & 44     & 41             & \textbf{85} \\ \hair
Environments          & 12     & 41             & \textbf{53} \\ \hair
\grp{Corpus per task} \\
Env files, mean       & 62.8   & 58.7           & \textbf{60.8} \\ \hair
Env files, max        & 112    & 207            & \textbf{207} \\ \hair
\grp{Grading} \\
Grader                & rubric & rubric + exec. & \textbf{hybrid} \\ \hair
Checks per task, mean & 21.4   & 29.9           & \textbf{25.3} \\[2pt]
\end{tabular}}%
\newlength{\statHh}\setlength{\statHh}{\dimexpr\ht\statHead+\dp\statHead\relax}%
\newlength{\statH}\setlength{\statH}{\dimexpr\statHh+\ht\statBody+\dp\statBody\relax}%
\newlength{\statW}\setlength{\statW}{\wd\statHead}%
\begin{minipage}[t]{0.485\linewidth}
\centering
\begin{tikzpicture}[x=1pt, y=1pt]
\begin{scope}
  \clip[rounded corners=6pt] (0,0) rectangle (\statW,-\statH);
  \fill[white] (0,0) rectangle (\statW,-\statH);
  \fill[webWash] (0,0) rectangle (\statW,-\statHh);
\end{scope}
\draw[webHairline, line width=0.4pt] (0,-\statHh) -- (\statW,-\statHh);
\node[inner sep=0pt, anchor=north west] at (0,0) {\usebox{\statHead}};
\node[inner sep=0pt, anchor=north west] at (0,-\statHh) {\usebox{\statBody}};
\draw[webHairline, line width=0.5pt, rounded corners=6pt] (0,0) rectangle (\statW,-\statH);
\end{tikzpicture}\par
\vspace{4pt}
\captionof{table}{\bench{} summary statistics. Tasks that share environment files count as one environment, while file counts are per task. Checks per task are averaged over the 81 tasks graded by rubric items or constraints (44 expert-authored and 37 co-authored).}
\label{tab:stats}
\end{minipage}\hfill
\begin{minipage}[t]{0.485\linewidth}
\centering
\pgfmathsetmacro{\Hc}{\statH/1cm}%
\pgfmathsetmacro{\Hh}{\statHh/1cm}%
\begin{tikzpicture}[x=1cm, y=1cm, font=\sffamily,
  cat/.style={font=\sffamily\footnotesize, text=webInk, anchor=east, inner sep=0pt},
  seg/.style={font=\sffamily\fontsize{6.6}{7.8}\selectfont\bfseries, inner sep=0pt},
  total/.style={font=\sffamily\fontsize{7.2}{8.5}\selectfont\bfseries, text=webInk, anchor=west, inner sep=0pt},
  key/.style={font=\sffamily\scriptsize, text=webInk, anchor=west, inner sep=0pt},
]
\pgfmathsetmacro{\xz}{1.62}\pgfmathsetmacro{\xs}{0.192}%
\pgfmathsetmacro{\ya}{0.42}
\pgfmathsetmacro{\yb}{-\Hc+\ya}%
\pgfmathsetmacro{\yt}{-\Hh-0.05}
\pgfmathsetmacro{\pitch}{(\yb-\yt)/4}
\pgfmathsetmacro{\bh}{0.42*abs(\pitch)}
\path (0,0) rectangle (8.0,-\Hc);                  
\fill[webBlue, rounded corners=1pt] (0.05,{-\Hh/2-0.11}) rectangle ++(0.22,0.22);
\node[key] at (0.36,-\Hh/2) {Expert-Authored \textcolor{webGraphite}{(44)}};
\fill[webBlueLight, rounded corners=1pt] (3.25,{-\Hh/2-0.11}) rectangle ++(0.22,0.22);
\node[key] at (3.56,-\Hh/2) {Co-Authored \textcolor{webGraphite}{(41)}};
\foreach \t in {0,5,...,30} {
  \draw[webHairline!55, line width=0.35pt] ({\xz+\t*\xs},\yb) -- ({\xz+\t*\xs},\yt);
  \node[font=\sffamily\scriptsize, text=webGraphite, anchor=north, inner sep=0pt] at ({\xz+\t*\xs},\yb-0.09) {\t};
}
\draw[webGraphite, line width=0.6pt] (\xz,\yb) -- ({\xz+30*\xs},\yb);
\node[font=\fontsize{6.2}{7}\selectfont\sffamily\bfseries, text=webGraphite, anchor=north east, inner sep=0pt]
  at ({\xz+30*\xs},\yb-0.36) {\textls[110]{TASKS}};
\foreach \name/\r/\e/\c in {Mechanical/0/19/11, Electrical/1/15/12, Civil/2/10/8, Structural/3/0/10} {
  \pgfmathsetmacro{\y}{\yt+(\r+0.5)*\pitch}
  \pgfmathsetmacro{\xe}{\xz+\e*\xs}
  \pgfmathsetmacro{\xc}{\xz+(\e+\c)*\xs}
  \node[cat] at (\xz-0.14,\y) {\name};
  \ifnum\e>0
    \fill[webBlue] (\xz,\y-\bh/2) rectangle (\xe,\y+\bh/2);
    \node[seg, text=white] at ({(\xz+\xe)/2},\y) {\e};
  \fi
  \fill[webBlueLight] (\xe,\y-\bh/2) rectangle (\xc,\y+\bh/2);
  \node[seg, text=webInk] at ({(\xe+\xc)/2},\y) {\c};
  \pgfmathtruncatemacro{\tot}{\e+\c}
  \node[total] at (\xc+0.1,\y) {\tot};
}
\end{tikzpicture}\par
\vspace{4pt}

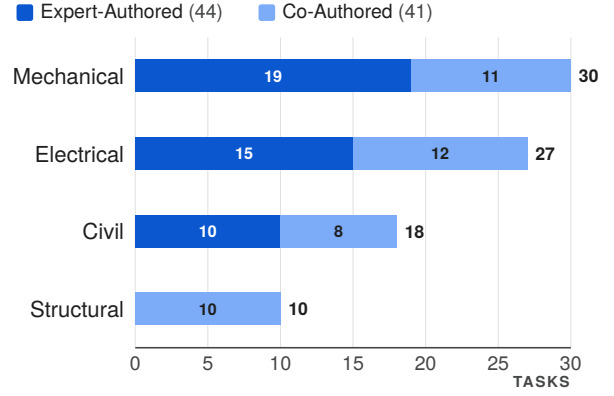
\captionof{figure}{Task distribution across physical-engineering disciplines and curation methods.}
\label{fig:disciplines}
\end{minipage}
\end{table}

\paragraph{The Creator--Solver loop.}
\Cref{fig:pipeline} (stages 2 and 3) shows this loop. The Creator drafts all three components together: the task statement, the environment, and an
executable grader with its reference deliverable. An independent \emph{Task Solver} then attacks the task by trying to solve it.
It makes $N$ parallel attempts (typically $N{=}3$) in sandboxed containers, with network access
restricted to the model endpoint and no access to the grader, rubrics or references. A task
meets the \emph{difficulty target} only if the \emph{mean} of the attempts stays below the
difficulty bar. The creator reads the solver's trajectories (which file it opened, which
shortcut or exploit it found). The creator checks a solved task and identifies the component that made the task easy. If a
component's difficulty can be intensified, the task undergoes another round of \emph{redesign} by modifying the
objective function or adding constraints to the grader. If the environment itself is too thin, the task returns to
deep research, which finds more diverse context documents to make the task more challenging.

\paragraph{Mercor validation and difficulty intensification.}
Every co-authored task is delivered to Mercor for independent expert validation. An expert
reviews the prompt, environment files, grader and reference deliverables, and confirms that they are realistic and
consistent with each other. A task that can be made
harder has its \emph{difficulty intensified}: the expert adds their own items and modifies the prompt, environment,
grader or reference deliverables accordingly.

\section{Data Analysis}
\label{sec:data_analysis}

\begin{figure}[!t]
\centering
\definecolor{webInk}{HTML}{1F1F1F}%
\definecolor{webGraphite}{HTML}{444746}%
\definecolor{webHairline}{HTML}{C4C7C5}%
\definecolor{webBlue}{HTML}{0B57D0}%
\definecolor{webBlueLight}{HTML}{7CACF8}%
\colorlet{webTrack}{webGraphite!18}%
\begin{tikzpicture}[x=1cm, y=1cm, font=\sffamily,
  tick/.style={anchor=east, inner sep=0pt, font=\sffamily\scriptsize, text=webGraphite},
  value/.style={anchor=south, inner sep=0pt, font=\sffamily\fontsize{5.9}{7}\selectfont, text=webInk},
  catlbl/.style={anchor=north, inner sep=0pt, font=\sffamily\fontsize{6.5}{7.6}\selectfont\bfseries, text=webInk},
  catsub/.style={anchor=north, inner sep=0pt, font=\sffamily\fontsize{5.6}{6.6}\selectfont, text=webGraphite},
  key/.style={anchor=west, inner sep=0pt, font=\sffamily\scriptsize, text=webInk},
  group/.style={anchor=south, inner sep=0pt, font=\sffamily\fontsize{6.4}{7.6}\selectfont, text=webGraphite},
  icon/.style={draw=webBlue, line width=0.55pt, line cap=round, line join=round},
]
\def\yscale{0.055}%
\foreach \p in {10,20,30,40,50,60} { \draw[webHairline!55, line width=0.35pt] (0,\p*\yscale) -- (14.8,\p*\yscale); }
\foreach \p in {0,10,20,30,40,50,60} { \node[tick] at (-0.12,\p*\yscale) {\p\%}; }
\draw[webGraphite, line width=0.6pt] (0,0) -- (14.8,0);
\node[anchor=south west, inner sep=0pt, font=\fontsize{6.2}{7}\selectfont\sffamily\bfseries, text=webGraphite]
  at (-0.62,4.04) {\textls[110]{SHARE OF ENV FILES (\%)}};
\node[key, anchor=east] (kc) at (14.8,4.15) {Co-Authored \textcolor{webGraphite}{(2,408 files, 41 tasks)}};
\fill[webBlueLight, rounded corners=0.8pt] ([xshift=-0.30cm,yshift=-0.1cm]kc.west) rectangle ++(0.2,0.2);
\node[key, anchor=east] (ke) at ([xshift=-0.75cm]kc.west) {Expert-Authored \textcolor{webGraphite}{(2,762 files, 44 tasks)}};
\fill[webBlue, rounded corners=0.8pt] ([xshift=-0.30cm,yshift=-0.1cm]ke.west) rectangle ++(0.2,0.2);
\draw[webHairline, line width=0.7pt] (0.30,3.42) -- (0.30,3.50) -- (5.34,3.50) -- (5.34,3.42);
\node[group] at (2.82,3.58) {\textbf{Documents, tables \& logs} (99.2\% vs.\ 75.0\%)};
\draw[webHairline, line width=0.7pt] (6.04,3.42) -- (6.04,3.50) -- (14.50,3.50) -- (14.50,3.42);
\node[group] at (10.27,3.58) {\textbf{CAD/BIM models, schematics, web \& code} (0.8\% vs.\ 25.0\%)};
\newcommand{\pair}[3]{%
  \pgfmathsetmacro{\hE}{max(#2*\yscale,0.035)}\pgfmathsetmacro{\hC}{max(#3*\yscale,0.035)}%
  \ifdim #2pt>0.01pt
    \fill[webBlue] (#1-0.405,0) rectangle (#1-0.025,\hE);
    \node[value] at (#1-0.215,\hE+0.035) {#2};
  \else
    \draw[webBlue!40, line width=0.8pt] (#1-0.37,0.02) -- (#1-0.06,0.02);
    \node[value, text=webGraphite] at (#1-0.215,0.06) {0.0};
  \fi
  \fill[webBlueLight] (#1+0.025,0) rectangle (#1+0.405,\hC);
  \node[value] at (#1+0.215,\hC+0.035) {#3};}
\pair{1.11}{51.6}{31.2}
\pair{2.82}{25.2}{27.5}
\pair{4.53}{22.4}{16.3}
\pair{6.84}{0.2}{12.9}
\pair{8.56}{0.3}{7.1}
\pair{10.27}{0.0}{2.5}
\pair{11.98}{0.0}{0.4}
\pair{13.69}{0.2}{2.0}
\begin{scope}[shift={(8.95,1.50)}]
  \draw[webHairline, line width=0.5pt, fill=white, rounded corners=4pt] (0,0) rectangle (5.6,1.36);
  \node[anchor=west, inner sep=0pt, font=\sffamily\fontsize{6.5}{7.6}\selectfont\bfseries, text=webInk] at (0.22,1.06)
    {Mean corpus disk footprint per task};
  \node[anchor=west, inner sep=0pt, font=\sffamily\fontsize{6.1}{7.2}\selectfont, text=webGraphite] at (0.22,0.66) {Expert-Authored};
  \fill[webTrack, rounded corners=0.6pt] (1.95,0.605) rectangle (3.85,0.715);
  \fill[webBlue, rounded corners=0.6pt] (1.95,0.605) rectangle (2.10,0.715);
  \node[anchor=west, inner sep=0pt, font=\sffamily\fontsize{6.1}{7.2}\selectfont\bfseries, text=webInk] at (3.98,0.66) {3.0\,MB};
  \node[anchor=west, inner sep=0pt, font=\sffamily\fontsize{6.1}{7.2}\selectfont, text=webGraphite] at (0.22,0.32) {Co-Authored};
  \fill[webBlueLight, rounded corners=0.6pt] (1.95,0.265) rectangle (3.85,0.375);
  \node[anchor=west, inner sep=0pt, font=\sffamily\fontsize{6.1}{7.2}\selectfont\bfseries, text=webInk] at (3.98,0.32)
    {38.1\,MB \mdseries\textcolor{webGraphite}{(12.6$\times$)}};
\end{scope}
\begin{scope}[shift={(1.11,-0.27)}]   
  \draw[icon] (-0.10,-0.13) -- (0.10,-0.13) -- (0.10,0.06) -- (0.03,0.13) -- (-0.10,0.13) -- cycle;
  \draw[icon, line width=0.45pt] (0.03,0.13) -- (0.03,0.06) -- (0.10,0.06) (-0.05,-0.01) -- (0.05,-0.01) (-0.05,-0.06) -- (0.03,-0.06);
\end{scope}
\node[catlbl] at (1.11,-0.50) {PDF docs};
\node[catsub] at (1.11,-0.76) {(.pdf)};
\begin{scope}[shift={(2.82,-0.27)}]   
  \draw[icon] (-0.13,-0.11) rectangle (0.13,0.11);
  \draw[icon, line width=0.45pt] (-0.13,0.035) -- (0.13,0.035) (-0.13,-0.04) -- (0.13,-0.04) (-0.03,-0.11) -- (-0.03,0.11);
\end{scope}
\node[catlbl] at (2.82,-0.50) {Spreadsheets};
\node[catsub] at (2.82,-0.76) {(.xlsx, .csv, .json)};
\begin{scope}[shift={(4.53,-0.27)}]   
  \draw[icon] (-0.11,-0.13) rectangle (0.11,0.13);
  \draw[icon, line width=0.45pt] (-0.06,0.06) -- (0.06,0.06) (-0.06,0.0) -- (0.06,0.0) (-0.06,-0.06) -- (0.02,-0.06);
\end{scope}
\node[catlbl] at (4.53,-0.50) {Text \& mail};
\node[catsub] at (4.53,-0.76) {(.docx, .eml, .md)};
\begin{scope}[shift={(6.84,-0.27)}]   
  \draw[icon] (0,0.135) -- (0.12,0.065) -- (0.12,-0.065) -- (0,-0.135) -- (-0.12,-0.065) -- (-0.12,0.065) -- cycle
    (-0.12,0.065) -- (0,-0.005) -- (0.12,0.065) (0,-0.005) -- (0,-0.135);
\end{scope}
\node[catlbl] at (6.84,-0.50) {CAD \& BIM};
\node[catsub] at (6.84,-0.76) {(.dxf, .dwg, .ifc)};
\begin{scope}[shift={(8.56,-0.27)}]   
  \draw[icon] (-0.10,-0.08) -- (0,0.08) -- (0.10,-0.08) (-0.05,0.0) -- (0.05,0.0);
  \fill[webBlue] (-0.10,-0.08) circle (1.1pt) (0,0.08) circle (1.1pt) (0.10,-0.08) circle (1.1pt);
\end{scope}
\node[catlbl] at (8.56,-0.50) {Schematics};
\node[catsub] at (8.56,-0.76) {(.png, .gif)};
\begin{scope}[shift={(10.27,-0.27)}]   
  \draw[icon] (0,0) circle (0.12);
  \draw[icon, line width=0.45pt] (0,0) ellipse (0.055 and 0.12) (-0.12,0) -- (0.12,0);
\end{scope}
\node[catlbl] at (10.27,-0.50) {Web portals};
\node[catsub] at (10.27,-0.76) {(.html)};
\begin{scope}[shift={(11.98,-0.27)}]   
  \draw[icon, line width=0.6pt] (-0.06,0.08) -- (-0.13,0.0) -- (-0.06,-0.08) (0.06,0.08) -- (0.13,0.0) -- (0.06,-0.08);
  \draw[icon, line width=0.5pt] (-0.02,-0.09) -- (0.02,0.09);
\end{scope}
\node[catlbl] at (11.98,-0.50) {Source code};
\node[catsub] at (11.98,-0.76) {(.py, .cc)};
\begin{scope}[shift={(13.69,-0.27)}]  
  \draw[icon, rounded corners=0.6pt] (-0.12,-0.10) -- (-0.12,0.10) -- (-0.03,0.10) -- (0.01,0.05) -- (0.12,0.05) -- (0.12,-0.10) -- cycle;
  \draw[icon, line width=0.45pt] (-0.12,-0.01) -- (0.12,-0.01);
\end{scope}
\node[catlbl] at (13.69,-0.50) {Other};
\node[catsub] at (13.69,-0.76) {(archives)};
\end{tikzpicture}
\caption{Environment file types by task curation approach. Expert-authored environments are
document-centric; co-authored environments add CAD/BIM models, schematics and imagery, web portals and code, and are
$12.6\times$ larger on disk per task.}
\label{fig:corpora}
\end{figure}
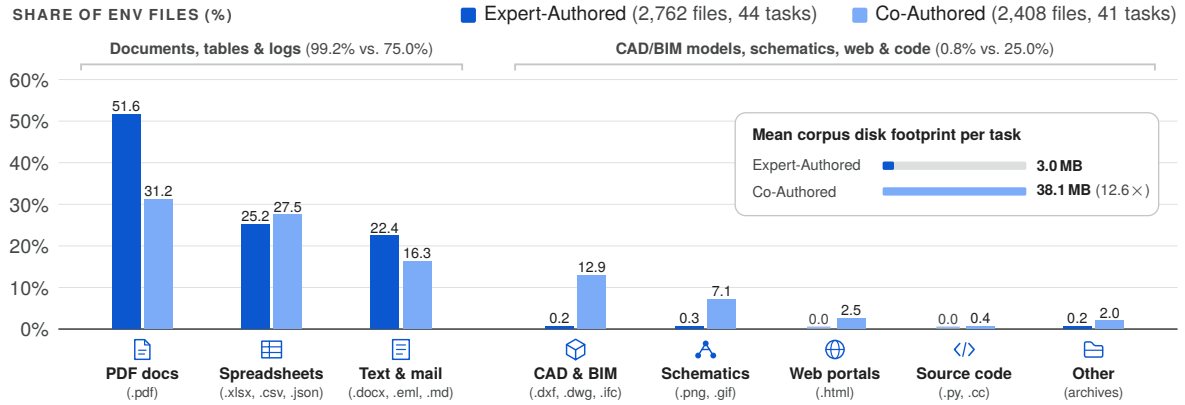

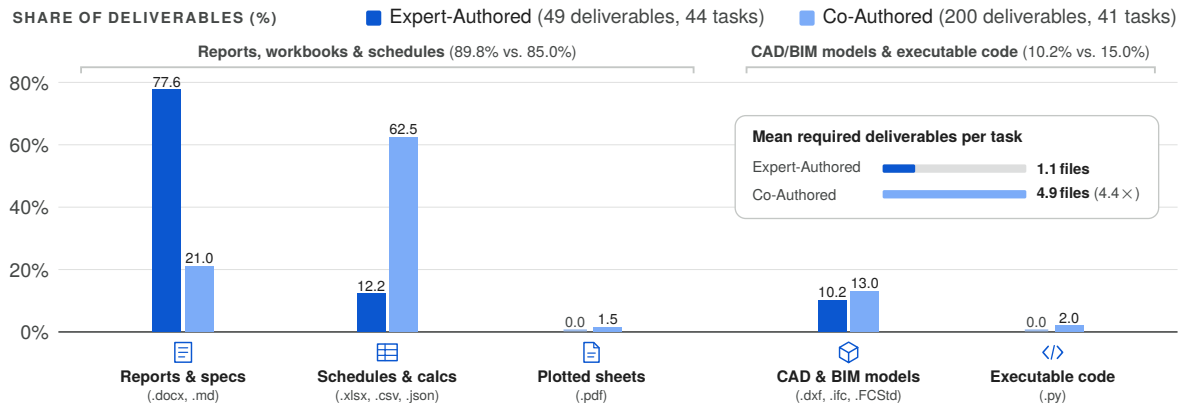
\begin{figure}[!t]
\centering
\definecolor{webInk}{HTML}{1F1F1F}%
\definecolor{webGraphite}{HTML}{444746}%
\definecolor{webHairline}{HTML}{C4C7C5}%
\definecolor{webBlue}{HTML}{0B57D0}%
\definecolor{webBlueLight}{HTML}{7CACF8}%
\colorlet{webTrack}{webGraphite!18}%
\begin{tikzpicture}[x=1cm, y=1cm, font=\sffamily,
  tick/.style={anchor=east, inner sep=0pt, font=\sffamily\scriptsize, text=webGraphite},
  value/.style={anchor=south, inner sep=0pt, font=\sffamily\fontsize{5.9}{7}\selectfont, text=webInk},
  catlbl/.style={anchor=north, inner sep=0pt, font=\sffamily\fontsize{6.5}{7.6}\selectfont\bfseries, text=webInk},
  catsub/.style={anchor=north, inner sep=0pt, font=\sffamily\fontsize{5.6}{6.6}\selectfont, text=webGraphite},
  key/.style={anchor=west, inner sep=0pt, font=\sffamily\scriptsize, text=webInk},
  group/.style={anchor=south, inner sep=0pt, font=\sffamily\fontsize{6.4}{7.6}\selectfont, text=webGraphite},
  icon/.style={draw=webBlue, line width=0.55pt, line cap=round, line join=round},
]
\def\yscale{0.04125}%
\foreach \p in {20,40,60,80} { \draw[webHairline!55, line width=0.35pt] (0,\p*\yscale) -- (14.8,\p*\yscale); }
\foreach \p in {0,20,40,60,80} { \node[tick] at (-0.12,\p*\yscale) {\p\%}; }
\draw[webGraphite, line width=0.6pt] (0,0) -- (14.8,0);
\node[anchor=south west, inner sep=0pt, font=\fontsize{6.2}{7}\selectfont\sffamily\bfseries, text=webGraphite]
  at (-0.62,4.04) {\textls[110]{SHARE OF DELIVERABLES (\%)}};
\node[key, anchor=east] (kc) at (14.8,4.15) {Co-Authored \textcolor{webGraphite}{(200 deliverables, 41 tasks)}};
\fill[webBlueLight, rounded corners=0.8pt] ([xshift=-0.30cm,yshift=-0.1cm]kc.west) rectangle ++(0.2,0.2);
\node[key, anchor=east] (ke) at ([xshift=-0.75cm]kc.west) {Expert-Authored \textcolor{webGraphite}{(49 deliverables, 44 tasks)}};
\fill[webBlue, rounded corners=0.8pt] ([xshift=-0.30cm,yshift=-0.1cm]ke.west) rectangle ++(0.2,0.2);
\draw[webHairline, line width=0.7pt] (0.30,3.42) -- (0.30,3.50) -- (8.40,3.50) -- (8.40,3.42);
\node[group] at (4.35,3.58) {\textbf{Reports, workbooks \& schedules} (89.8\% vs.\ 85.0\%)};
\draw[webHairline, line width=0.7pt] (9.10,3.42) -- (9.10,3.50) -- (14.50,3.50) -- (14.50,3.42);
\node[group] at (11.80,3.58) {\textbf{CAD/BIM models \& executable code} (10.2\% vs.\ 15.0\%)};
\newcommand{\pair}[3]{%
  \pgfmathsetmacro{\hE}{max(#2*\yscale,0.035)}\pgfmathsetmacro{\hC}{max(#3*\yscale,0.035)}%
  \ifdim #2pt>0.01pt
    \fill[webBlue] (#1-0.405,0) rectangle (#1-0.025,\hE);
    \node[value] at (#1-0.215,\hE+0.035) {#2};
  \else
    \draw[webBlue!40, line width=0.8pt] (#1-0.37,0.02) -- (#1-0.06,0.02);
    \node[value, text=webGraphite] at (#1-0.215,0.06) {0.0};
  \fi
  \fill[webBlueLight] (#1+0.025,0) rectangle (#1+0.405,\hC);
  \node[value] at (#1+0.215,\hC+0.035) {#3};}
\pair{1.65}{77.6}{21.0}
\pair{4.35}{12.2}{62.5}
\pair{7.05}{0.0}{1.5}
\pair{10.45}{10.2}{13.0}
\pair{13.15}{0.0}{2.0}
\begin{scope}[shift={(8.95,1.50)}]
  \draw[webHairline, line width=0.5pt, fill=white, rounded corners=4pt] (0,0) rectangle (5.6,1.36);
  \node[anchor=west, inner sep=0pt, font=\sffamily\fontsize{6.5}{7.6}\selectfont\bfseries, text=webInk] at (0.22,1.06)
    {Mean required deliverables per task};
  \node[anchor=west, inner sep=0pt, font=\sffamily\fontsize{6.1}{7.2}\selectfont, text=webGraphite] at (0.22,0.66) {Expert-Authored};
  \fill[webTrack, rounded corners=0.6pt] (1.95,0.605) rectangle (3.85,0.715);
  \fill[webBlue, rounded corners=0.6pt] (1.95,0.605) rectangle (2.38,0.715);
  \node[anchor=west, inner sep=0pt, font=\sffamily\fontsize{6.1}{7.2}\selectfont\bfseries, text=webInk] at (3.98,0.66) {1.1\,files};
  \node[anchor=west, inner sep=0pt, font=\sffamily\fontsize{6.1}{7.2}\selectfont, text=webGraphite] at (0.22,0.32) {Co-Authored};
  \fill[webBlueLight, rounded corners=0.6pt] (1.95,0.265) rectangle (3.85,0.375);
  \node[anchor=west, inner sep=0pt, font=\sffamily\fontsize{6.1}{7.2}\selectfont\bfseries, text=webInk] at (3.98,0.32)
    {4.9\,files \mdseries\textcolor{webGraphite}{(4.4$\times$)}};
\end{scope}
\begin{scope}[shift={(1.65,-0.27)}]   
  \draw[icon] (-0.11,-0.13) rectangle (0.11,0.13);
  \draw[icon, line width=0.45pt] (-0.06,0.06) -- (0.06,0.06) (-0.06,0.0) -- (0.06,0.0) (-0.06,-0.06) -- (0.02,-0.06);
\end{scope}
\node[catlbl] at (1.65,-0.50) {Reports \& specs};
\node[catsub] at (1.65,-0.76) {(.docx, .md)};
\begin{scope}[shift={(4.35,-0.27)}]   
  \draw[icon] (-0.13,-0.11) rectangle (0.13,0.11);
  \draw[icon, line width=0.45pt] (-0.13,0.035) -- (0.13,0.035) (-0.13,-0.04) -- (0.13,-0.04) (-0.03,-0.11) -- (-0.03,0.11);
\end{scope}
\node[catlbl] at (4.35,-0.50) {Schedules \& calcs};
\node[catsub] at (4.35,-0.76) {(.xlsx, .csv, .json)};
\begin{scope}[shift={(7.05,-0.27)}]   
  \draw[icon] (-0.10,-0.13) -- (0.10,-0.13) -- (0.10,0.06) -- (0.03,0.13) -- (-0.10,0.13) -- cycle;
  \draw[icon, line width=0.45pt] (0.03,0.13) -- (0.03,0.06) -- (0.10,0.06) (-0.05,-0.01) -- (0.05,-0.01) (-0.05,-0.06) -- (0.03,-0.06);
\end{scope}
\node[catlbl] at (7.05,-0.50) {Plotted sheets};
\node[catsub] at (7.05,-0.76) {(.pdf)};
\begin{scope}[shift={(10.45,-0.27)}]  
  \draw[icon] (0,0.135) -- (0.12,0.065) -- (0.12,-0.065) -- (0,-0.135) -- (-0.12,-0.065) -- (-0.12,0.065) -- cycle
    (-0.12,0.065) -- (0,-0.005) -- (0.12,0.065) (0,-0.005) -- (0,-0.135);
\end{scope}
\node[catlbl] at (10.45,-0.50) {CAD \& BIM models};
\node[catsub] at (10.45,-0.76) {(.dxf, .ifc, .FCStd)};
\begin{scope}[shift={(13.15,-0.27)}]  
  \draw[icon, line width=0.6pt] (-0.06,0.08) -- (-0.13,0.0) -- (-0.06,-0.08) (0.06,0.08) -- (0.13,0.0) -- (0.06,-0.08);
  \draw[icon, line width=0.5pt] (-0.02,-0.09) -- (0.02,0.09);
\end{scope}
\node[catlbl] at (13.15,-0.50) {Executable code};
\node[catsub] at (13.15,-0.76) {(.py)};
\end{tikzpicture}
\caption{Deliverable file type by curation approach (249 deliverables across 85 tasks). Expert-authored tasks
mostly issue one formal report or workbook; co-authored tasks issue multi-file packages.}
\label{fig:deliverables}
\end{figure}

\Cref{tab:stats} summarises \bench{}. The two curation approaches contribute 44 expert-authored and 41
co-authored tasks set in 53 distinct environments: the expert-authored tasks share 12 project environments, while
every co-authored task has its own. Each environment is a corpus, not a prompt attachment. In total the agent faces
5,170 environment files, 61 per task on average and up to 207. Expert-authored tasks are graded against 941 rubric items
(21.4 per task), and the 37 constraint-graded co-authored tasks against 29.9 constraints per task on average, on top of their
objective score (25.3 checks per task overall). Every task asks for one or more deliverable files, and none can be answered with a simple LLM call.

The two approaches complement each other in discipline (\cref{fig:disciplines}). Mechanical (30 tasks) and
electrical (27) engineering are the largest disciplines, followed by civil (18) and structural (10). The
expert-authored tasks are concentrated in mechanical, electrical and civil practice, while all ten structural tasks,
and the building-systems design work in the other disciplines, come from the co-authored route.

The two approaches also stress different skills on the input and output sides. Expert-authored environments are
document-centric (\cref{fig:corpora}): 51.6\% PDFs, 25.2\% spreadsheets and data tables, and 22.4\% text, Word
documents and e-mail. Co-authored environments balance PDFs (31.2\%), spreadsheets and data (27.5\%) and text
(16.3\%) with CAD and BIM models (12.9\%), schematics and imagery (7.1\%), web portals (2.5\%) and code, and are an
order of magnitude larger on disk (38.1\,MB per task against 3.0\,MB). Both include distractors such as superseded
revisions and unrelated sheets, so the agent must decide which document governs before it computes anything. On the
output side (\cref{fig:deliverables}), the 85 tasks require 249 deliverable files. Expert-authored tasks mostly issue
a single formal report (\texttt{.docx}, 77.6\%), workbook (12.2\%) or CAD model (10.2\%), 1.1 files per task, whereas
co-authored tasks issue multi-file engineering packages of 4.9 files on average: schedules, calculation
workbooks and data files (62.5\%), design narratives and specifications (21.0\%), 2D CAD drawings and 3D BIM models
(13.0\%), executable scripts (2.0\%) and plotted sheets (1.5\%).

\section{Results}
\label{sec:results}

\subsection{Experimental Setup}
\label{sec:setup}

\paragraph{Agent harnesses.}
We evaluate 14 models, \modelset{}, each under seven publicly available agent harnesses (GPT~6~Astra only in Codex and Grok~4.7 only in its own harness, Grok Build~\citep{xai2026grokbuild}): Antigravity~\citep{google2025antigravity}, Claude Code~\citep{anthropic2025claudecode}, Codex~\citep{openai2025codex}, Hermes~\citep{nous2025hermesagent}, OpenCode~\citep{opencode2025}, OpenHands~\citep{wang2024openhands} and
Pi~\citep{zechner2025pi}. Crossing models with harnesses separates how much of the score
depends on the model and how much on the scaffolding around it. Each harness runs inside the task's container with the
environment mounted. Docker-environment tasks expose their applications (filesystem, documents,
spreadsheets, PDFs, mail, calendar, chat) as MCP tool servers; Pi has no MCP support and reaches
them through the shell. During the agent phase the container can reach only a model proxy, which
pins one reasoning effort for every harness; every run has a 6-hour agent budget. When the agent
exits, its deliverable files are collected and graded outside the agent's container, either
host-side or in a separate verifier container that receives only the task's declared artifacts,
so the solver can never reach the grader or the reference deliverables.
Harness versions and settings are listed in \cref{app:config}.

\paragraph{Grading.}
The agent sees only the instruction and the environment, with no rubric, verifier or score feedback.
Rubric items of the expert-authored tasks are judged by Claude Opus~5.5, which reads the
deliverable file itself. Co-authored tasks are graded by their frozen executable graders and reported with \cref{eq:score}; twelve of these graders
also ask Claude Opus~5.5 to judge three to six narrative items, which count as soft constraints in $p$. The two site-design and two MEP-coordination tasks keep their graders' own scores, which give 0\% to an MEP routing with
a hard clash or an unserved row and at most 50\% to an infeasible site design.
Task scores lie in $[0,1]$ and we report them in percent. Every model runs every
task three times in each of its harnesses, and we report the mean. A run that fails for infrastructure reasons (e.g.\ a
provider error before the first model call) is re-run; a run that ends without a gradable
deliverable scores 0\%. All results are over the 85 tasks of \bench{} (\cref{tab:main}).

\subsection{Main Results}

\begin{table}[t]
\centering
\caption{Results of Claude Opus~5.5 in each of the seven agent harnesses on all 85 \bench{} tasks. All numbers are
scores in percent (\%), the mean task score over three runs (higher is better; best harness per column in bold).
Rubric items are judged by Claude Opus~5.5, and co-authored tasks are scored with \cref{eq:score}. Task counts in
parentheses.}
\label{tab:main}
\definecolor{webInk}{HTML}{1F1F1F}%
\definecolor{webGraphite}{HTML}{444746}%
\definecolor{webHairline}{HTML}{C4C7C5}%
\definecolor{webWash}{HTML}{F0F4F9}%
\definecolor{webBlue}{HTML}{0B57D0}%
\colorlet{webTint}{webBlue!13}%
\footnotesize\sffamily\color{webInk}%
\newcommand{\thd}[1]{\textcolor{webGraphite}{\scriptsize\bfseries #1}}%
\newcommand{\tsub}[1]{\textcolor{webGraphite}{\scriptsize #1}}%
\newcommand{\grp}[1]{\textcolor{webGraphite}{\scriptsize\bfseries #1}}%
\newcommand{\best}[1]{\textcolor{webBlue}{\textbf{#1}}}%
\newcommand{\hair}{\arrayrulecolor{webHairline}\specialrule{0.4pt}{0pt}{0pt}}%
\newlength{\mainH}\settowidth{\mainH}{\bfseries Claude Code}%
\newlength{\mainK}\setlength{\mainK}{\dimexpr(\textwidth-0.5pt-\mainH-50pt)/7\relax}%
\newcolumntype{H}{>{\raggedright\arraybackslash}p{\mainH}}%
\newcolumntype{K}{>{\centering\arraybackslash}p{\mainK}}%
\setlength{\tabcolsep}{0pt}%
\def\maincols{@{\hspace{12pt}}HK@{\hspace{12pt}}KK@{\hspace{12pt}}KKKK@{\hspace{14pt}}}%
\newsavebox{\mainHead}\newsavebox{\mainBody}\newsavebox{\mainRow}%
\sbox{\mainHead}{\renewcommand{\arraystretch}{1.0}\expandafter\tabular\expandafter{\maincols}
\rule{0pt}{3.0ex} & & \multicolumn{2}{c}{\grp{Curation approach}} & \multicolumn{4}{c}{\grp{Discipline}} \\
\arrayrulecolor{webHairline}\cmidrule[0.4pt](lr{6pt}){3-4}\cmidrule[0.4pt](lr{6pt}){5-8}
\rule{0pt}{2.2ex}\thd{Harness} & \thd{All} & \thd{Expert} & \thd{Co-authored} & \thd{Electrical} & \thd{Civil} & \thd{Mechanical} & \thd{Structural} \\
\rule[-1.5ex]{0pt}{2.4ex} & \tsub{(85)} & \tsub{(44)} & \tsub{(41)} & \tsub{(27)} & \tsub{(18)} & \tsub{(30)} & \tsub{(10)} \\
\endtabular}%
\sbox{\mainBody}{\renewcommand{\arraystretch}{1.42}\expandafter\tabular\expandafter{\maincols}
Pi & 47.3 & 49.2 & 45.3 & 46.7 & 60.7 & 40.5 & 45.0 \\ \hair
OpenCode & 47.4 & 50.6 & 44.1 & 47.9 & 62.1 & 39.2 & 44.4 \\ \hair
Antigravity & 48.4 & 50.8 & 45.8 & 49.5 & 62.1 & \best{41.4} & 41.6 \\ \hair
Hermes & 48.6 & 49.2 & 48.0 & 51.9 & 65.1 & 38.4 & 40.8 \\ \hair
Codex & 49.3 & 50.6 & 48.0 & \best{53.1} & 67.8 & 36.6 & 43.8 \\ \hair
OpenHands & 49.9 & 50.7 & \best{49.1} & 50.6 & 67.0 & 40.2 & \best{46.6} \\
\endtabular}%
\sbox{\mainRow}{\renewcommand{\arraystretch}{1.0}\expandafter\tabular\expandafter{\maincols}
\rule[-1.25ex]{0pt}{3.6ex}\textbf{Claude Code} & \best{50.6} & \best{53.1} & 48.0 & 50.7 & \best{69.9} & 40.5 & \best{46.6} \\
\endtabular}%
\newlength{\mainHh}\setlength{\mainHh}{\dimexpr\ht\mainHead+\dp\mainHead\relax}%
\newlength{\mainBh}\setlength{\mainBh}{\dimexpr\ht\mainBody+\dp\mainBody\relax}%
\newlength{\mainRh}\setlength{\mainRh}{\dimexpr\ht\mainRow+\dp\mainRow\relax}%
\newlength{\mainW}\setlength{\mainW}{\wd\mainHead}%
\begin{tikzpicture}[x=1pt, y=1pt]
\begin{scope}
  \clip[rounded corners=6pt] (0,0) rectangle (\mainW,{-\mainHh-\mainBh-\mainRh-6pt});
  \fill[white] (0,0) rectangle (\mainW,{-\mainHh-\mainBh-\mainRh-6pt});
  \fill[webWash] (0,0) rectangle (\mainW,-\mainHh);
\end{scope}
\draw[webHairline, line width=0.4pt] (0,-\mainHh) -- (\mainW,-\mainHh);
\fill[webTint, rounded corners={0.5\mainRh}] (4pt,{-\mainHh-\mainBh-2pt}) rectangle (\mainW-4pt,{-\mainHh-\mainBh-2pt-\mainRh});
\node[inner sep=0pt, anchor=north west] at (0,0) {\usebox{\mainHead}};
\node[inner sep=0pt, anchor=north west] at (0,-\mainHh) {\usebox{\mainBody}};
\node[inner sep=0pt, anchor=north west] at (0,{-\mainHh-\mainBh-2pt}) {\usebox{\mainRow}};
\draw[webHairline, line width=0.5pt, rounded corners=6pt] (0,0) rectangle (\mainW,{-\mainHh-\mainBh-\mainRh-6pt});
\end{tikzpicture}%
\arrayrulecolor{pleHairline}
\end{table}

\begin{figure}[t]
\centering
\definecolor{webInk}{HTML}{1F1F1F}%
\definecolor{webGraphite}{HTML}{444746}%
\definecolor{webHairline}{HTML}{C4C7C5}%
\definecolor{webBlue}{HTML}{0B57D0}%
\definecolor{webBlueLight}{HTML}{7CACF8}%
\definecolor{webRed}{HTML}{D93025}%
\begin{tikzpicture}
\begin{axis}[
  scale only axis, width=0.92\linewidth, height=3.3cm,
  ybar, bar width=3.3pt, bar shift=0pt,
  ymin=0, ymax=100, xmin=0, xmax=86, clip=false,
  axis lines*=left, x axis line style={draw=webGraphite, line width=0.6pt, -}, y axis line style={draw=none},
  xtick=\empty, ytick={0,25,50,75,100}, yticklabels={0\%,25\%,50\%,75\%,100\%}, ytick style={draw=none},
  ymajorgrids, major grid style={draw=webHairline!55, line width=0.35pt},
  tick label style={font=\scriptsize\sffamily, text=webGraphite}, yticklabel style={xshift=-1pt},
  legend style={font=\scriptsize\sffamily, text=webInk, draw=none, fill=none, legend columns=-1,
    at={(0,1)}, anchor=south west, yshift=17pt, xshift=-28pt,
    /tikz/every even column/.append style={column sep=8pt}},
  legend cell align=left,
]
\addplot[fill=webBlue, draw=none, legend image code/.code={\fill[webBlue, rounded corners=0.8pt] (0pt,-2.7pt) rectangle (5.4pt,2.7pt);}]
  table[x=rank, y=score] {data/fig_pertask_expert.dat};
\addplot[fill=webBlueLight, draw=none, legend image code/.code={\fill[webBlueLight, rounded corners=0.8pt] (0pt,-2.7pt) rectangle (5.4pt,2.7pt);}]
  table[x=rank, y=score] {data/fig_pertask_coauth.dat};
\addplot[webRed, line width=0.9pt, dash pattern=on 3pt off 2pt, sharp plot, forget plot] coordinates {(0,50.6) (86,50.6)};
\node[anchor=south west, font=\scriptsize\sffamily\bfseries, text=webRed, fill=white, inner sep=1.5pt] at (axis cs:1,52.0) {Mean: 50.6\%};
\legend{Expert-Authored, Co-Authored}
\node[font=\fontsize{6.2}{7}\selectfont\sffamily\bfseries, text=webGraphite, anchor=south west, inner sep=0pt] at ([shift={(-28pt,7pt)}]rel axis cs:0,1) {\textls[110]{SCORE (\%)}};
\node[font=\fontsize{6.2}{7}\selectfont\sffamily\bfseries, text=webGraphite, anchor=north east, inner sep=0pt] at ([yshift=-6pt]rel axis cs:1,0) {\textls[110]{TASKS, SORTED BY SCORE (CLAUDE OPUS 5.5 IN CLAUDE CODE)}};
\end{axis}
\end{tikzpicture}
\caption{Per-task scores on all 85 tasks with Claude Opus~5.5 in Claude Code
(mean of three runs). The dashed redline indicates the mean score (50.6\%).}
\label{fig:pertask}
\end{figure}
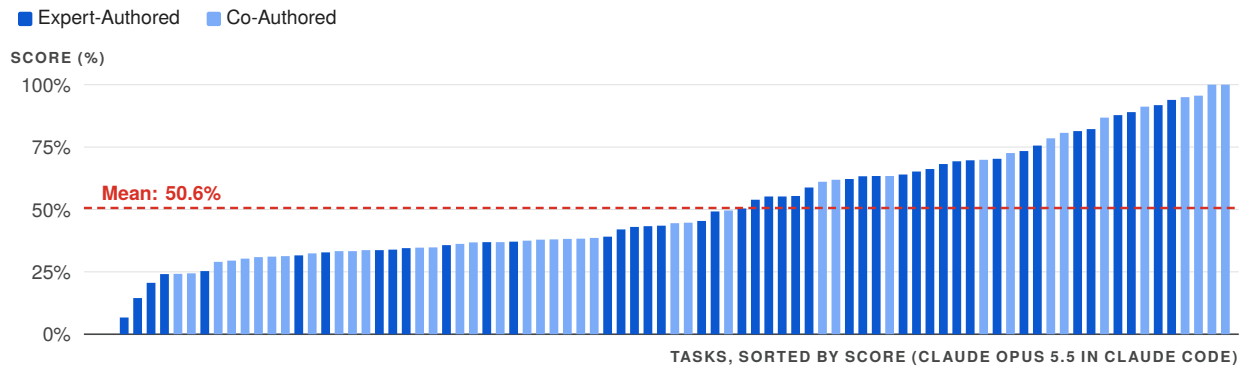

\Cref{fig:pareto} places all 86 model (harness) pairs by score and cost, \cref{tab:main} breaks down the
strongest model by harness, curation approach and discipline, \cref{fig:pertask} shows its per-task
distribution, and \cref{tab:model_harness} lists every pair.

\paragraph{Performance and cost.}
Three proprietary model families hold the upper performance-cost frontier (above \$2 per task, as shown in \cref{fig:pareto}). At the top, Claude Opus~5.5 scores 50.6\% at \$17.18 per task in Claude Code and 49.9\% at \$14.82 in OpenHands, while GPT~6~Astra in Codex comes within 2.3 points of the top score at about half the cost (48.3\%, \$9.15). Gemini~4~Argon covers the \$6--9 band across four harnesses (40.2\%--43.2\%), and Gemini~3.8~Flash in Claude Code retains 56\% of the top score for 15\% of its cost (28.2\%, \$2.55). Below \$0.20 per task, the open-weight models on the frontier score under 7\%. Higher cost does not always yield a higher score: the eight most expensive configurations, all Claude models in Antigravity, Claude Code or Codex, cost \$27--54 per task yet score only 33.8\%--49.3\%, all below Claude Opus~5.5 in Claude Code (\$17.18 per task).

\paragraph{\bench{} is far from saturated.}
The strongest configuration, Claude Opus~5.5 in Claude Code, averages 50.6\% (\cref{tab:model_harness}), ahead of the best configurations of the next two models: GPT~6~Astra in Codex (48.3\%) and Claude Sonnet~5.5 in Antigravity (45.2\%). In Claude Code (\cref{tab:main}), Claude Opus~5.5 has a median task score of 43.5\% and scores below 50\% on 48 of the 85 tasks; even taking the best harness separately for each task leaves 35 tasks below 50\%. Two tasks score 0\% in every run, and seven score above 90\%. By discipline, mechanical (40.5\%) and structural (46.6\%) work are the hardest, while civil is the easiest (69.9\%). Co-authored tasks are harder than expert-authored ones in every harness (48.0\% versus 53.1\% in Claude Code).

\paragraph{The gap between models.}
The model matters far more than the harness (\cref{tab:model_harness}). Taking each model's best harness, scores fall from 50.6\% (Claude Opus~5.5) and 48.3\% (GPT~6~Astra) to 45.2\% (Claude Sonnet~5.5), 43.2\% (Gemini~4~Argon) and 42.6\% (Claude Opus~5). They then drop to 34.9\% for Claude Fable~5, 32.0\% for Grok~4.7, 28.2\% for Gemini~3.8~Flash, 25.3\% for GLM~5.2 and 14.2\% for Gemini~3.1~Pro, while Qwen3~Coder~480B, MiniMax~M2, Kimi~K2~Thinking and Llama~4~Maverick stay below 7\%. This 50-point spread across models far exceeds the 0.7--12.1 point spread across harnesses for a fixed model (median 3.0). Costs vary just as widely (\cref{fig:pareto}): the best configurations of Claude Opus~5.5 and GPT~6~Astra cost \$17.18 and \$9.15 per task, while Gemini~4~Argon reaches 43.2\% for \$8.21 and Gemini~3.8~Flash reaches 28.2\% for \$2.55.

\paragraph{The harness's effect.}
For Claude Opus~5.5, the seven harnesses span only 47.3\%--50.6\%, much less than the spread across tasks (\cref{fig:pertask}). Harness rankings also vary across models (\cref{tab:model_harness}). Claude Code is the best harness for Claude Opus~5.5, Gemini~4~Argon, Claude Fable~5 and Gemini~3.8~Flash; Antigravity for Claude Sonnet~5.5 and Claude Opus~5; Hermes for Gemini~3.1~Pro and Qwen3~Coder~480B; Codex for GLM~5.2; and OpenCode for Kimi~K2~Thinking. MiniMax~M2 (Claude Code and Hermes) and Llama~4~Maverick (OpenHands and Pi) each tie between two harnesses. Runs that score 0\% contribute to these differences; they end without a gradable deliverable or fail all-or-nothing gates such as those of site design and MEP coordination. Out of 255 runs per harness, Claude Opus~5.5 has 16 such runs in Pi, 15 in Codex, 12 in OpenCode, 11 in Claude Code, 8 in Hermes, 7 in Antigravity and 6 in OpenHands.

\subsection{Analysis}
\label{sec:analysis}

\paragraph{Run-to-run variance is substantial for single tasks.}
Set-level means are stable: for Claude Opus~5.5, each harness's mean has a standard deviation of 0.3--1.8 points across its three runs (e.g.\ Claude Code 52.0\% / 49.3\% / 50.6\%). Individual tasks, by contrast, vary widely. The mean range of a task-harness cell over three runs is 15.4 points, and 105 of the 595 cells span 30 points or more. Reported per-task numbers should therefore always be averages over repeated runs.

\paragraph{Co-authored tasks demand far more work.}
For Claude Opus~5.5 in Claude Code, an attempt at a co-authored task consumes 39.3M tokens and 97 minutes of agent time, compared with 15.6M tokens and 31 minutes for an expert-authored task (27.0M tokens and 63 minutes overall). Harnesses also differ widely in inference cost at similar scores (\cref{fig:pareto}). For Claude Opus~5.5, Claude Code (\$17.18/task, 50.6\%) and OpenHands (\$14.82/task, 49.9\%) sit on the performance-cost frontier, while OpenCode (\$12.53/task, 47.4\%) and Pi (\$13.33/task, 47.3\%) come close at lower cost; Codex, by contrast, averages \$35.15/task (49.3\%).

\paragraph{Effect of the rubric judge.}
We re-graded all 1,159 LLM-judged runs of Claude Opus~5.5 across the seven harnesses using a second pass of Claude Opus~5.5, Claude Opus~5, and Gemini~4~Argon. The second Claude Opus~5.5 pass reproduces 98.7\% of item verdicts (Cohen's $\kappa=0.97$) with an unchanged mean. Against it, Claude Opus~5 and Gemini~4~Argon agree on 95.8\% and 95.5\% of items ($\kappa=0.90$), raising re-graded scores by 3.3 and 1.3 points (2.1 and 0.9 overall), almost entirely on rubric-graded tasks. Claude Code ranks first under every judge: Claude Opus~5 preserves the harness order (Kendall's $\tau=1.0$), and Gemini~4~Argon swaps only Codex and Hermes ($\tau=0.90$). On the 44 rubric-graded tasks in Claude Code, switching to Gemini~4~Argon raises its score by 3.6 points versus 1.3 for Claude Opus~5.5, a 2.3-point combined self-preference.

\paragraph{Where agents fail.}
Across all runs of the 86 model (harness) pairs, 23.2\% score 0\%. Of these, 25.3\% end without a gradable deliverable, 4.3\% hit the agent's time budget or an agent-side error (a crashed harness, an exhausted context window or an output limit), and 70.4\% deliver files that the grader scores 0\%. This last group comprises 2,020 runs on tasks scored with \cref{eq:score}, 817 on rubric-graded tasks and 726 at the all-or-nothing gates of the site-design and MEP-coordination graders. The strongest configuration, Claude Opus~5.5 in Claude Code, rarely fails outright, scoring 0\% in only 11 of its 255 runs. Its losses stem instead from hard constraints: on co-authored tasks, its deliverable items satisfy 91\% of their constraints on average, yet only 41.3\% of the items satisfy every hard constraint and so earn objective credit under \cref{eq:score}. Producing a design that is fully safe and feasible while pushing its cost toward the bound remains the core difficulty of \bench{}.

\section{Conclusion}
\label{sec:conclusion}

We introduced \bench{}, a benchmark of 85 agentic tasks that evaluates whether frontier models can perform the
realistic, economically valuable work of professional physical engineers. Solving a task requires an agent to
navigate a multi-format project corpus, reconcile the governing physical and code requirements, and issue
production-ready engineering deliverables. The tasks come from two complementary curation approaches: expert-authored
tasks written end to end by practising engineers and graded with weighted rubrics, and co-authored tasks built from
engineers' briefs by an adversarial Creator--Solver agent pipeline and then validated and intensified by experts. For
design tasks, a hybrid score rewards a cost-effective, fully compliant design rather than a merely admissible one.

Across 14 frontier models and seven agent harnesses, only one configuration exceeds 50\%: Claude Opus~5.5 in Claude Code, at 50.6\%. The model matters far more than the harness, yet the harness still shifts scores and costs substantially; results on \bench{} should therefore always be reported with both the model and the harness. The main obstacle remains producing a design that satisfies every hard constraint while still optimising its cost, leaving substantial headroom for future agents.

\paragraph{Ethical discussion.}
No task contains proprietary or client-confidential information: expert-authored environments are written from
scratch by the participating engineers, and every file in a co-authored environment comes from a public-domain or
permissively licensed source, or is generated for the task, under a licence that permits redistribution. The
engineers were paid through Mercor at professional rates: on average \$4,663 per expert-authored task and \$3,157
per co-authored task for validation and difficulty intensification.

\section*{Acknowledgements}
\label{sec:ack}
We are grateful to Michael Colesanti, Dan Roush, Jeff Thiele, Amanda Gunckle, Julie Sellers, Daniela Zellers, Xia
Fang, Patricia Perez, and Shilpa Seroha, who helped review and contributed to this work.

\bibliography{references}

@article{glazer2024frontiermath,
  title={{FrontierMath: A Benchmark for Evaluating Advanced Mathematical Reasoning in AI}},
  author={Elliot Glazer and Ege Erdil and Tamay Besiroglu and Diego Chicharro and Evan Chen and Alex Gunning and Caroline Falkman Olsson and Jean-Stanislas Denain and Anson Ho and Emily de Oliveira Santos and others},
  journal={arXiv preprint arXiv:2411.04872},
  year={2024},
  eprint={2411.04872},
  archivePrefix={arXiv},
  url={https://arxiv.org/abs/2411.04872}
}

@article{jimenez2023swebench,
  title={{SWE-bench: Can Language Models Resolve Real-World GitHub Issues?}},
  author={Carlos E. Jimenez and John Yang and Alexander Wettig and Shunyu Yao and Kexin Pei and Ofir Press and Karthik Narasimhan},
  journal={arXiv preprint arXiv:2310.06770},
  year={2023},
  eprint={2310.06770},
  archivePrefix={arXiv},
  url={https://arxiv.org/abs/2310.06770}
}

@article{yang2024swebenchmm,
  title={{SWE-bench Multimodal: Do AI Systems Generalize to Visual Software Domains?}},
  author={John Yang and Carlos E. Jimenez and Alex L. Zhang and Kilian Lieret and Joyce Yang and Xindi Wu and Ori Press and Niklas Muennighoff and Gabriel Synnaeve and Karthik R. Narasimhan and others},
  journal={arXiv preprint arXiv:2410.03859},
  year={2024},
  eprint={2410.03859},
  archivePrefix={arXiv},
  url={https://arxiv.org/abs/2410.03859}
}

@article{miserendino2025swelancer,
  title={{SWE-Lancer: Can Frontier LLMs Earn \$1 Million from Real-World Freelance Software Engineering?}},
  author={Samuel Miserendino and Michele Wang and Tejal Patwardhan and Johannes Heidecke},
  journal={arXiv preprint arXiv:2502.12115},
  year={2025},
  eprint={2502.12115},
  archivePrefix={arXiv},
  url={https://arxiv.org/abs/2502.12115}
}

@article{chan2024mlebench,
  title={{MLE-bench: Evaluating Machine Learning Agents on Machine Learning Engineering}},
  author={Jun Shern Chan and Neil Chowdhury and Oliver Jaffe and James Aung and Dane Sherburn and Evan Mays and Giulio Starace and Kevin Liu and Leon Maksin and Tejal Patwardhan and others},
  journal={arXiv preprint arXiv:2410.07095},
  year={2024},
  eprint={2410.07095},
  archivePrefix={arXiv},
  url={https://arxiv.org/abs/2410.07095}
}

@article{wijk2024rebench,
  title={{RE-Bench: Evaluating frontier AI R\&D capabilities of language model agents against human experts}},
  author={Hjalmar Wijk and Tao Lin and Joel Becker and Sami Jawhar and Neev Parikh and Thomas Broadley and Lawrence Chan and Michael Chen and Josh Clymer and Jai Dhyani and others},
  journal={arXiv preprint arXiv:2411.15114},
  year={2024},
  eprint={2411.15114},
  archivePrefix={arXiv},
  url={https://arxiv.org/abs/2411.15114}
}

@article{starace2025paperbench,
  title={{PaperBench: Evaluating AI's Ability to Replicate AI Research}},
  author={Giulio Starace and Oliver Jaffe and Dane Sherburn and James Aung and Jun Shern Chan and Leon Maksin and Rachel Dias and Evan Mays and Benjamin Kinsella and Wyatt Thompson and others},
  journal={arXiv preprint arXiv:2504.01848},
  year={2025},
  eprint={2504.01848},
  archivePrefix={arXiv},
  url={https://arxiv.org/abs/2504.01848}
}

@article{merrill2026terminalbench,
  title={{Terminal-Bench: Benchmarking Agents on Hard, Realistic Tasks in Command Line Interfaces}},
  author={Mike A. Merrill and Alexander G. Shaw and Nicholas Carlini and Boxuan Li and Harsh Raj and Ivan Bercovich and Lin Shi and Jeong Yeon Shin and Thomas Walshe and E. Kelly Buchanan and others},
  journal={arXiv preprint arXiv:2601.11868},
  year={2026},
  eprint={2601.11868},
  archivePrefix={arXiv},
  url={https://arxiv.org/abs/2601.11868}
}

@article{xie2024osworld,
  title={{OSWorld: Benchmarking Multimodal Agents for Open-Ended Tasks in Real Computer Environments}},
  author={Tianbao Xie and Danyang Zhang and Jixuan Chen and Xiaochuan Li and Siheng Zhao and Ruisheng Cao and Toh Jing Hua and Zhoujun Cheng and Dongchan Shin and Fangyu Lei and others},
  journal={arXiv preprint arXiv:2404.07972},
  year={2024},
  eprint={2404.07972},
  archivePrefix={arXiv},
  url={https://arxiv.org/abs/2404.07972}
}

@article{zhou2023webarena,
  title={{WebArena: A Realistic Web Environment for Building Autonomous Agents}},
  author={Shuyan Zhou and Frank F. Xu and Hao Zhu and Xuhui Zhou and Robert Lo and Abishek Sridhar and Xianyi Cheng and Tianyue Ou and Yonatan Bisk and Daniel Fried and others},
  journal={arXiv preprint arXiv:2307.13854},
  year={2023},
  eprint={2307.13854},
  archivePrefix={arXiv},
  url={https://arxiv.org/abs/2307.13854}
}

@article{mialon2023gaia,
  title={{GAIA: a benchmark for General AI Assistants}},
  author={Grégoire Mialon and Clémentine Fourrier and Craig Swift and Thomas Wolf and Yann LeCun and Thomas Scialom},
  journal={arXiv preprint arXiv:2311.12983},
  year={2023},
  eprint={2311.12983},
  archivePrefix={arXiv},
  url={https://arxiv.org/abs/2311.12983}
}

@article{xu2024theagentcompany,
  title={{TheAgentCompany: Benchmarking LLM Agents on Consequential Real World Tasks}},
  author={Frank F. Xu and Yufan Song and Boxuan Li and Yuxuan Tang and Kritanjali Jain and Mengxue Bao and Zora Z. Wang and Xuhui Zhou and Zhitong Guo and Murong Cao and others},
  journal={arXiv preprint arXiv:2412.14161},
  year={2024},
  eprint={2412.14161},
  archivePrefix={arXiv},
  url={https://arxiv.org/abs/2412.14161}
}

@article{yao2024taubench,
  title={{$\tau$-bench: A Benchmark for Tool-Agent-User Interaction in Real-World Domains}},
  author={Shunyu Yao and Noah Shinn and Pedram Razavi and Karthik Narasimhan},
  journal={arXiv preprint arXiv:2406.12045},
  year={2024},
  eprint={2406.12045},
  archivePrefix={arXiv},
  url={https://arxiv.org/abs/2406.12045}
}

@article{zhang2024cybench,
  title={{Cybench: A Framework for Evaluating Cybersecurity Capabilities and Risks of Language Models}},
  author={Andy K. Zhang and Neil Perry and Riya Dulepet and Joey Ji and Celeste Menders and Justin W. Lin and Eliot Jones and Gashon Hussein and Samantha Liu and Donovan Jasper and others},
  journal={arXiv preprint arXiv:2408.08926},
  year={2024},
  eprint={2408.08926},
  archivePrefix={arXiv},
  url={https://arxiv.org/abs/2408.08926}
}

@article{tian2024scicode,
  title={{SciCode: A Research Coding Benchmark Curated by Scientists}},
  author={Minyang Tian and Luyu Gao and Shizhuo Dylan Zhang and Xinan Chen and Cunwei Fan and Xuefei Guo and Roland Haas and Pan Ji and Kittithat Krongchon and Yao Li and others},
  journal={arXiv preprint arXiv:2407.13168},
  year={2024},
  eprint={2407.13168},
  archivePrefix={arXiv},
  url={https://arxiv.org/abs/2407.13168}
}

@article{patwardhan2025gdpval,
  title={{GDPval: Evaluating AI Model Performance on Real-World Economically Valuable Tasks}},
  author={Tejal Patwardhan and Rachel Dias and Elizabeth Proehl and Grace Kim and Michele Wang and Olivia Watkins and Simón Posada Fishman and Marwan Aljubeh and Phoebe Thacker and Laurance Fauconnet and others},
  journal={arXiv preprint arXiv:2510.04374},
  year={2025},
  eprint={2510.04374},
  archivePrefix={arXiv},
  url={https://arxiv.org/abs/2510.04374}
}

@article{vidgen2025apex,
  title={{The AI Productivity Index (APEX)}},
  author={Bertie Vidgen and Abby Fennelly and Evan Pinnix and Julien Benchek and Daniyal Khan and Zach Richards and Austin Bridges and Calix Huang and Kanishka Sahu and Abhishek Kottamasu and others},
  journal={arXiv preprint arXiv:2509.25721},
  year={2025},
  eprint={2509.25721},
  archivePrefix={arXiv},
  url={https://arxiv.org/abs/2509.25721}
}

@article{vidgen2026apexagents,
  title={{APEX-Agents}},
  author={Bertie Vidgen and Austin Mann and Abby Fennelly and John Wright Stanly and Lucas Rothman and Marco Burstein and Julien Benchek and David Ostrofsky and Anirudh Ravichandran and Debnil Sur and others},
  journal={arXiv preprint arXiv:2601.14242},
  year={2026},
  eprint={2601.14242},
  archivePrefix={arXiv},
  url={https://arxiv.org/abs/2601.14242}
}

@article{arora2025healthbench,
  title={{HealthBench: Evaluating Large Language Models Towards Improved Human Health}},
  author={Rahul K. Arora and Jason Wei and Rebecca Soskin Hicks and Preston Bowman and Joaquin Quiñonero-Candela and Foivos Tsimpourlas and Michael Sharman and Meghan Shah and Andrea Vallone and Alex Beutel and others},
  journal={arXiv preprint arXiv:2505.08775},
  year={2025},
  eprint={2505.08775},
  archivePrefix={arXiv},
  url={https://arxiv.org/abs/2505.08775}
}

@article{guo2025engdesign,
  title={{Toward Engineering AGI: Benchmarking the Engineering Design Capabilities of LLMs}},
  author={Xingang Guo and Yaxin Li and Xiangyi Kong and Yilan Jiang and Xiayu Zhao and Zhihua Gong and Yufan Zhang and Daixuan Li and Tianle Sang and Beixiao Zhu and others},
  journal={arXiv preprint arXiv:2509.16204},
  year={2025},
  eprint={2509.16204},
  archivePrefix={arXiv},
  url={https://arxiv.org/abs/2509.16204}
}

@article{zhou2025engibench,
  title={{EngiBench: A Benchmark for Evaluating Large Language Models on Engineering Problem Solving}},
  author={Xiyuan Zhou and Xinlei Wang and Yirui He and Yang Wu and Ruixi Zou and Yuheng Cheng and Yulu Xie and Wenxuan Liu and Huan Zhao and Yan Xu and others},
  journal={arXiv preprint arXiv:2509.17677},
  year={2025},
  eprint={2509.17677},
  archivePrefix={arXiv},
  url={https://arxiv.org/abs/2509.17677}
}

@article{chi2026frontiereng,
  title={{Frontier-Eng: Benchmarking Self-Evolving Agents on Real-World Engineering Tasks with Generative Optimization}},
  author={Yizhe Chi and Deyao Hong and Dapeng Jiang and Tianwei Luo and Kaisen Yang and Boshi Zhang and Zhe Cao and Xiaoyan Fan and Bingxiang He and Han Hao and others},
  journal={arXiv preprint arXiv:2604.12290},
  year={2026},
  eprint={2604.12290},
  archivePrefix={arXiv},
  url={https://arxiv.org/abs/2604.12290}
}

@article{mudur2025feabench,
  title={{FEABench: Evaluating Language Models on Multiphysics Reasoning Ability}},
  author={Nayantara Mudur and Hao Cui and Subhashini Venugopalan and Paul Raccuglia and Michael P. Brenner and Peter Norgaard},
  journal={arXiv preprint arXiv:2504.06260},
  year={2025},
  eprint={2504.06260},
  archivePrefix={arXiv},
  url={https://arxiv.org/abs/2504.06260}
}

@article{doris2024designqa,
  title={{DesignQA: A Multimodal Benchmark for Evaluating Large Language Models' Understanding of Engineering Documentation}},
  author={Anna C. Doris and Daniele Grandi and Ryan Tomich and Md Ferdous Alam and Mohammadmehdi Ataei and Hyunmin Cheong and Faez Ahmed},
  journal={arXiv preprint arXiv:2404.07917},
  year={2024},
  eprint={2404.07917},
  archivePrefix={arXiv},
  url={https://arxiv.org/abs/2404.07917}
}

@article{doris2026cadbench,
  title={{CADBench: A Multimodal Benchmark for AI-Assisted CAD Program Generation}},
  author={Anna C. Doris and Jacob Thomas Sony and Ghadi Nehme and Era Syla and Amin Heyrani Nobari and Faez Ahmed},
  journal={arXiv preprint arXiv:2605.10873},
  year={2026},
  eprint={2605.10873},
  archivePrefix={arXiv},
  url={https://arxiv.org/abs/2605.10873}
}

@article{khan2024text2cad,
  title={{Text2CAD: Generating Sequential CAD Models from Beginner-to-Expert Level Text Prompts}},
  author={Mohammad Sadil Khan and Sankalp Sinha and Talha Uddin Sheikh and Didier Stricker and Sk Aziz Ali and Muhammad Zeshan Afzal},
  journal={arXiv preprint arXiv:2409.17106},
  year={2024},
  eprint={2409.17106},
  archivePrefix={arXiv},
  url={https://arxiv.org/abs/2409.17106}
}

@article{lai2024analogcoder,
  title={{AnalogCoder: Analog Circuit Design via Training-Free Code Generation}},
  author={Yao Lai and Sungyoung Lee and Guojin Chen and Souradip Poddar and Mengkang Hu and David Z. Pan and Ping Luo},
  journal={arXiv preprint arXiv:2405.14918},
  year={2024},
  eprint={2405.14918},
  archivePrefix={arXiv},
  url={https://arxiv.org/abs/2405.14918}
}

@article{bhandari2024masalachai,
  title={{Masala-CHAI: A Large-Scale SPICE Netlist Dataset for Analog Circuits by Harnessing AI}},
  author={Jitendra Bhandari and Vineet Bhat and Yuheng He and Hamed Rahmani and Siddharth Garg and Ramesh Karri},
  journal={arXiv preprint arXiv:2411.14299},
  year={2024},
  eprint={2411.14299},
  archivePrefix={arXiv},
  url={https://arxiv.org/abs/2411.14299}
}

@article{zhou2024elecbench,
  title={{ElecBench: a Power Dispatch Evaluation Benchmark for Large Language Models}},
  author={Xiyuan Zhou and Huan Zhao and Yuheng Cheng and Yuji Cao and Gaoqi Liang and Guolong Liu and Wenxuan Liu and Yan Xu and Junhua Zhao},
  journal={arXiv preprint arXiv:2407.05365},
  year={2024},
  eprint={2407.05365},
  archivePrefix={arXiv},
  url={https://arxiv.org/abs/2407.05365}
}

@article{li2025drafterbench,
  title={{DrafterBench: Benchmarking Large Language Models for Tasks Automation in Civil Engineering}},
  author={Yinsheng Li and Zhen Dong and Yi Shao},
  journal={arXiv preprint arXiv:2507.11527},
  year={2025},
  eprint={2507.11527},
  archivePrefix={arXiv},
  url={https://arxiv.org/abs/2507.11527}
}

@article{liang2025masse,
  title={{Automating Structural Engineering Workflows with Large Language Model Agents}},
  author={Haoran Liang and Yufa Zhou and Mohammad Talebi Kalaleh and Qipei Mei},
  journal={arXiv preprint arXiv:2510.11004},
  year={2025},
  eprint={2510.11004},
  archivePrefix={arXiv},
  url={https://arxiv.org/abs/2510.11004}
}

@article{neubauer2026hvacllm,
  title={{Large Language Models for HVAC Operations in Building Energy Systems: A Critical Review of Methods, Applications, and Deployment Readiness}},
  author={Alexander Neubauer and Tianzhen Hong and Han Li and Mengbo Yu and Amin Darbandi and Yannick Fürst and Martin Kriegel},
  journal={arXiv preprint arXiv:2609.05314},
  year={2026},
  eprint={2609.05314},
  archivePrefix={arXiv},
  url={https://arxiv.org/abs/2609.05314}
}

@article{zheng2023judging,
  title={{Judging LLM-as-a-Judge with MT-Bench and Chatbot Arena}},
  author={Lianmin Zheng and Wei-Lin Chiang and Ying Sheng and Siyuan Zhuang and Zhanghao Wu and Yonghao Zhuang and Zi Lin and Zhuohan Li and Dacheng Li and Eric P. Xing and others},
  journal={arXiv preprint arXiv:2306.05685},
  year={2023},
  eprint={2306.05685},
  archivePrefix={arXiv},
  url={https://arxiv.org/abs/2306.05685}
}

@article{kim2023prometheus,
  title={{Prometheus: Inducing Fine-grained Evaluation Capability in Language Models}},
  author={Seungone Kim and Jamin Shin and Yejin Cho and Joel Jang and Shayne Longpre and Hwaran Lee and Sangdoo Yun and Seongjin Shin and Sungdong Kim and James Thorne and others},
  journal={arXiv preprint arXiv:2310.08491},
  year={2023},
  eprint={2310.08491},
  archivePrefix={arXiv},
  url={https://arxiv.org/abs/2310.08491}
}

@article{gu2024llmjudgesurvey,
  title={{A Survey on LLM-as-a-Judge}},
  author={Jiawei Gu and Xuhui Jiang and Zhichao Shi and Hexiang Tan and Xuehao Zhai and Chengjin Xu and Wei Li and Yinghan Shen and Shengjie Ma and Honghao Liu and others},
  journal={arXiv preprint arXiv:2411.15594},
  year={2024},
  eprint={2411.15594},
  archivePrefix={arXiv},
  url={https://arxiv.org/abs/2411.15594}
}

@article{xu2024contamination,
  title={{Benchmark Data Contamination of Large Language Models: A Survey}},
  author={Cheng Xu and Shuhao Guan and Derek Greene and M-Tahar Kechadi},
  journal={arXiv preprint arXiv:2406.04244},
  year={2024},
  eprint={2406.04244},
  archivePrefix={arXiv},
  url={https://arxiv.org/abs/2406.04244}
}

@article{akhtar2026saturation,
  title={{When AI Benchmarks Plateau: A Systematic Study of Benchmark Saturation}},
  author={Mubashara Akhtar and Anka Reuel and Prajna Soni and Sanchit Ahuja and Pawan Sasanka Ammanamanchi and Ruchit Rawal and Vilém Zouhar and Srishti Yadav and Chenxi Whitehouse and Dayeon Ki and others},
  journal={arXiv preprint arXiv:2602.16763},
  year={2026},
  eprint={2602.16763},
  archivePrefix={arXiv},
  url={https://arxiv.org/abs/2602.16763}
}

@article{wang2024codeact,
  title={{Executable Code Actions Elicit Better LLM Agents}},
  author={Xingyao Wang and Yangyi Chen and Lifan Yuan and Yizhe Zhang and Yunzhu Li and Hao Peng and Heng Ji},
  journal={arXiv preprint arXiv:2402.01030},
  year={2024},
  eprint={2402.01030},
  archivePrefix={arXiv},
  url={https://arxiv.org/abs/2402.01030}
}

@article{yao2022react,
  title={{ReAct: Synergizing Reasoning and Acting in Language Models}},
  author={Shunyu Yao and Jeffrey Zhao and Dian Yu and Nan Du and Izhak Shafran and Karthik Narasimhan and Yuan Cao},
  journal={arXiv preprint arXiv:2210.03629},
  year={2022},
  eprint={2210.03629},
  archivePrefix={arXiv},
  url={https://arxiv.org/abs/2210.03629}
}

@article{wang2024openhands,
  title={{OpenHands: An Open Platform for AI Software Developers as Generalist Agents}},
  author={Xingyao Wang and Boxuan Li and Yufan Song and Frank F. Xu and Xiangru Tang and Mingchen Zhuge and Jiayi Pan and Yueqi Song and Bowen Li and Jaskirat Singh and others},
  journal={arXiv preprint arXiv:2407.16741},
  year={2024},
  eprint={2407.16741},
  archivePrefix={arXiv},
  url={https://arxiv.org/abs/2407.16741}
}

@misc{openai2024swebenchverified,
  title={{Introducing SWE-bench Verified}},
  author={{OpenAI}},
  year={2024},
  month=aug,
  howpublished={Blog post},
  url={https://openai.com/index/introducing-swe-bench-verified/}
}

@misc{mercor2026archipelago,
  title={{Archipelago}: Harness for running and evaluating {AI} agents against {RL} environments},
  author={{Mercor}},
  year={2026},
  howpublished={\url{https://github.com/Mercor-Intelligence/archipelago}},
  note={Software repository, Apache-2.0; introduced with APEX-Agents (arXiv:2601.14242)}
}

@techreport{asce2025reportcard,
  title={2025 {Report Card for America's Infrastructure}},
  author={{American Society of Civil Engineers}},
  institution={American Society of Civil Engineers},
  address={Reston, VA},
  year={2025},
  url={https://infrastructurereportcard.org/}
}

@article{xu2026sage,
  title={{{SAGE}: Steerable Agentic Data Generation for Deep Search with Execution Feedback}},
  author={Fangyuan Xu and Rujun Han and Yanfei Chen and Zifeng Wang and I-Hung Hsu and Jun Yan and Vishy Tirumalashetty and Eunsol Choi and Tomas Pfister and Chen-Yu Lee},
  journal={arXiv preprint arXiv:2601.18202},
  year={2026},
  eprint={2601.18202},
  archivePrefix={arXiv},
  url={https://arxiv.org/abs/2601.18202}
}

@techreport{gihub2017outlook,
  title={Global Infrastructure Outlook: Infrastructure Investment Needs, 50 Countries, 7 Sectors to 2040},
  author={{Global Infrastructure Hub and Oxford Economics}},
  institution={Global Infrastructure Hub},
  address={Sydney},
  year={2017},
  url={https://outlook.gihub.org/}
}

@misc{census2022sas,
  title={Service Annual Survey: Engineering Services ({NAICS} 54133), Total Revenue for Employer Firms (2022 data)},
  author={{U.S. Census Bureau}},
  year={2024},
  url={https://www.census.gov/programs-surveys/sas.html}
}

@misc{nous2025hermesagent,
  title={{Hermes Agent}},
  author={{Nous Research}},
  year={2025},
  howpublished={\url{https://github.com/NousResearch/hermes-agent}}
}

@misc{anthropic2025claudecode,
  title={{Claude Code}},
  author={{Anthropic}},
  year={2025},
  howpublished={\url{https://github.com/anthropics/claude-code}}
}

@misc{openai2025codex,
  title={Codex {CLI}},
  author={{OpenAI}},
  year={2025},
  howpublished={\url{https://github.com/openai/codex}}
}

@misc{opencode2025,
  title={{OpenCode}},
  author={{Anomaly}},
  year={2025},
  howpublished={\url{https://opencode.ai}}
}

@misc{zechner2025pi,
  title={{Pi}: a minimal terminal coding agent},
  author={Zechner, Mario},
  year={2025},
  howpublished={\url{https://github.com/badlogic/pi-mono}}
}

@misc{google2025antigravity,
  title={{Google Antigravity}},
  author={{Google}},
  year={2025},
  howpublished={\url{https://antigravity.google}}
}

@misc{xai2026grokbuild,
  title={{Grok Build}},
  author={{xAI}},
  year={2026},
  howpublished={\url{https://x.ai/cli}}
}

\appendix
\section{Evaluation Configuration}
\label{app:config}

\Cref{tab:config} lists the settings behind every result in \cref{sec:results}. All 14 models are served through
one model proxy, the only endpoint the agent container can reach, which pins each model's highest reasoning setting
in every harness. Each model and harness pair runs every task three times with a 6-hour agent budget per run. After
the agent exits, Claude Opus~5.5 judges the rubric items and co-authored tasks are scored by their frozen executable
graders outside the agent container (\cref{sec:setup}).

\begin{table}[h!]
\centering
\caption{Evaluation configuration for all results in \cref{sec:results}.}
\label{tab:config}
\definecolor{webInk}{HTML}{1F1F1F}%
\definecolor{webGraphite}{HTML}{444746}%
\definecolor{webHairline}{HTML}{C4C7C5}%
\definecolor{webWash}{HTML}{F0F4F9}%
\footnotesize\sffamily\color{webInk}%
\newlength{\cfgS}\settowidth{\cfgS}{\bfseries Runs per task, model \& harness}%
\newlength{\cfgV}\setlength{\cfgV}{\dimexpr\textwidth-\cfgS-36pt\relax}%
\newcolumntype{S}{>{\raggedright\arraybackslash}p{\cfgS}}%
\newcolumntype{V}{>{\raggedright\arraybackslash}p{\cfgV}}%
\setlength{\tabcolsep}{0pt}%
\newcommand{\thd}[1]{\textcolor{webGraphite}{\scriptsize\bfseries #1}}%
\newcommand{\hair}{\arrayrulecolor{webHairline}\specialrule{0.4pt}{0pt}{0pt}}%
\newsavebox{\cfgHead}\newsavebox{\cfgBody}%
\sbox{\cfgHead}{\begin{tabular}{@{\hspace{12pt}}S@{\hspace{12pt}}V@{\hspace{12pt}}}
\rule[-1.4ex]{0pt}{3.8ex}\thd{Setting} & \thd{Value} \\
\end{tabular}}%
\sbox{\cfgBody}{\renewcommand{\arraystretch}{1.45}\begin{tabular}{@{\hspace{12pt}}S@{\hspace{12pt}}V@{\hspace{12pt}}}
\textbf{Agent models} & 14 models: \texttt{claude-opus-5-5}, \texttt{claude-opus-5}, \texttt{claude-sonnet-5-5}, \texttt{claude-fable-5}, \texttt{gpt-6-astra}, \texttt{gemini-4-argon}, \texttt{gemini-3.8-flash}, \texttt{gemini-3.1-pro-preview}, \texttt{grok-4.7}, \texttt{glm-5.2}, \texttt{qwen3-coder-480b}, \texttt{minimax-m2}, \texttt{kimi-k2-thinking}, \texttt{llama-4-maverick}; all served through one model proxy \\ \hair
\textbf{Harnesses} & Antigravity 1.3.3, Claude Code 2.1.296, Codex 0.162.1, Hermes 0.22.2, OpenCode 2.0.26, OpenHands 1.54.0, Pi 1.1.0, Grok Build 1.0.52 \\ \hair
\textbf{Reasoning effort} & each model's highest reasoning setting (Qwen3 Coder 480B and Llama 4 Maverick have none) \\ \hair
\textbf{Network during agent phase} & no-internet \\ \hair
\textbf{Agent budget} & 6 hours (21{,}600\,s) of agent time per run \\ \hair
\textbf{Runs per task, model \& harness} & 3 \\ \hair
\textbf{Rubric judge} & \texttt{claude-opus-5-5} for every model, reads the deliverable file \\ \hair
\textbf{Co-authored grading} & frozen executable grader, \cref{eq:score}, outside the agent container \\
\end{tabular}}%
\newlength{\cfgHh}\setlength{\cfgHh}{\dimexpr\ht\cfgHead+\dp\cfgHead\relax}%
\newlength{\cfgBh}\setlength{\cfgBh}{\dimexpr\ht\cfgBody+\dp\cfgBody\relax}%
\newlength{\cfgW}\setlength{\cfgW}{\wd\cfgHead}%
\begin{tikzpicture}[x=1pt, y=1pt]
\begin{scope}
  \clip[rounded corners=6pt] (0,0) rectangle (\cfgW,{-\cfgHh-\cfgBh-2pt});
  \fill[white] (0,0) rectangle (\cfgW,{-\cfgHh-\cfgBh-2pt});
  \fill[webWash] (0,0) rectangle (\cfgW,-\cfgHh);
\end{scope}
\draw[webHairline, line width=0.4pt] (0,-\cfgHh) -- (\cfgW,-\cfgHh);
\node[inner sep=0pt, anchor=north west] at (0,0) {\usebox{\cfgHead}};
\node[inner sep=0pt, anchor=north west] at (0,-\cfgHh) {\usebox{\cfgBody}};
\draw[webHairline, line width=0.5pt, rounded corners=6pt] (0,0) rectangle (\cfgW,{-\cfgHh-\cfgBh-2pt});
\end{tikzpicture}%
\arrayrulecolor{pleHairline}
\end{table}
\FloatBarrier

\section{Results for Every Model and Harness}
\label{app:model_harness}

\Cref{tab:model_harness} gives the mean score of every model and harness pair on all 85 tasks; the model and harness
comparisons in \cref{sec:results} are drawn from it.

\begin{table}[h!]
\centering
\caption{Mean score in percent of every model (harness) pair (three runs each) on all 85 tasks
(best harness per model in bold). Rubric items are judged by Claude Opus~5.5, and co-authored tasks are scored with
\cref{eq:score}. \Cref{tab:main} breaks down the
Claude Opus~5.5 row by curation approach and discipline.}
\label{tab:model_harness}
\definecolor{webInk}{HTML}{1F1F1F}%
\definecolor{webGraphite}{HTML}{444746}%
\definecolor{webHairline}{HTML}{C4C7C5}%
\definecolor{webWash}{HTML}{F0F4F9}%
\definecolor{webBlue}{HTML}{0B57D0}%
\definecolor{webNo}{HTML}{8E918F}%
\footnotesize\sffamily\color{webInk}%
\newsavebox{\mhNo}\sbox{\mhNo}{\makebox[\widthof{00.0}]{\tikz[baseline=0pt]\draw[webNo, line width=0.8pt, line cap=round]
  (0,2.6pt) -- (5.2pt,2.6pt);}}%
\def\N{\usebox{\mhNo}}%
\newlength{\mhM}\settowidth{\mhM}{Qwen3 Coder 480B}%
\newcolumntype{M}{>{\raggedright\arraybackslash}p{\mhM}}%
\newcolumntype{H}{>{\raggedleft\arraybackslash}p{31pt}}%
\setlength{\tabcolsep}{0pt}%
\newcommand{\thd}[1]{\textcolor{webGraphite}{\scriptsize\bfseries #1}}%
\newcommand{\hair}{\arrayrulecolor{webHairline}\specialrule{0.4pt}{0pt}{0pt}}%
\newsavebox{\mhHead}\newsavebox{\mhBody}%
\sbox{\mhHead}{\begin{tabular}{@{\hspace{12pt}}M*{7}{H}@{\hspace{12pt}}}
\rule[-1.4ex]{0pt}{3.8ex}\thd{Model} & \thd{Agy} & \thd{CC} & \thd{Cdx} & \thd{Hrm} & \thd{OC} & \thd{OH} & \thd{Pi} \\
\end{tabular}}%
\sbox{\mhBody}{\renewcommand{\arraystretch}{1.24}\begin{tabular}{@{\hspace{12pt}}M*{7}{H}@{\hspace{12pt}}}
Claude Opus 5.5        & 48.4 & \textcolor{webBlue}{\textbf{50.6}} & 49.3 & 48.6 & 47.4 & 49.9 & 47.3 \\ \hair
GPT 6 Astra            & \N & \N & \textcolor{webBlue}{\textbf{48.3}} & \N & \N & \N & \N \\ \hair
Claude Sonnet 5.5      & \textcolor{webBlue}{\textbf{45.2}} & 43.6 & 42.7 & 43.7 & 42.3 & 42.2 & 43.5 \\ \hair
Gemini 4 Argon         & 41.7 & \textcolor{webBlue}{\textbf{43.2}} & 40.3 & 41.1 & 38.4 & 31.1 & 40.2 \\ \hair
Claude Opus 5          & \textcolor{webBlue}{\textbf{42.6}} & 40.8 & 38.8 & 39.6 & 39.8 & 41.9 & 40.4 \\ \hair
Claude Fable 5         & 33.8 & \textcolor{webBlue}{\textbf{34.9}} & 34.7 & 32.8 & 34.5 & 33.4 & 33.0 \\ \hair
Grok 4.7$^{\dagger}$ & \multicolumn{7}{c}{32.0 in Grok Build} \\ \hair
Gemini 3.8 Flash       & 26.4 & \textcolor{webBlue}{\textbf{28.2}} & 27.6 & 17.8 & 26.0 & 27.6 & 26.9 \\ \hair
GLM 5.2                & 23.0 & 22.3 & \textcolor{webBlue}{\textbf{25.3}} & 23.9 & 24.4 & 22.9 & 23.8 \\ \hair
Gemini 3.1 Pro         & 13.7 & 14.1 & 14.1 & \textcolor{webBlue}{\textbf{14.2}} & 13.3 & 14.0 & 13.6 \\ \hair
Qwen3 Coder 480B       & 4.5 & 5.6 & 4.4 & \textcolor{webBlue}{\textbf{6.2}} & 4.1 & 5.7 & 4.3 \\ \hair
Kimi K2 Thinking       & 3.5 & 4.4 & 3.0 & 5.2 & \textcolor{webBlue}{\textbf{5.9}} & 4.7 & 4.2 \\ \hair
MiniMax M2             & 3.7 & \textcolor{webBlue}{\textbf{5.3}} & 4.4 & \textcolor{webBlue}{\textbf{5.3}} & 3.9 & 4.5 & 4.7 \\ \hair
Llama 4 Maverick       & 0.1 & 0.4 & 0.0 & 0.1 & 0.0 & \textcolor{webBlue}{\textbf{0.7}} & \textcolor{webBlue}{\textbf{0.7}} \\
\end{tabular}}%
\newlength{\mhHh}\setlength{\mhHh}{\dimexpr\ht\mhHead+\dp\mhHead\relax}%
\newlength{\mhBh}\setlength{\mhBh}{\dimexpr\ht\mhBody+\dp\mhBody\relax}%
\newlength{\mhW}\setlength{\mhW}{\wd\mhHead}%
\begin{tikzpicture}[x=1pt, y=1pt]
\begin{scope}
  \clip[rounded corners=6pt] (0,0) rectangle (\mhW,{-\mhHh-\mhBh-2pt});
  \fill[white] (0,0) rectangle (\mhW,{-\mhHh-\mhBh-2pt});
  \fill[webWash] (0,0) rectangle (\mhW,-\mhHh);
\end{scope}
\draw[webHairline, line width=0.4pt] (0,-\mhHh) -- (\mhW,-\mhHh);
\node[inner sep=0pt, anchor=north west] at (0,0) {\usebox{\mhHead}};
\node[inner sep=0pt, anchor=north west] at (0,-\mhHh) {\usebox{\mhBody}};
\draw[webHairline, line width=0.5pt, rounded corners=6pt] (0,0) rectangle (\mhW,{-\mhHh-\mhBh-2pt});
\node[inner sep=0pt, anchor=north west, text width=\mhW, align=left, font=\scriptsize\sffamily, text=webGraphite]
  at (0,{-\mhHh-\mhBh-7pt}) {Agy~=~Antigravity; CC~=~Claude Code; Cdx~=~Codex; Hrm~=~Hermes; OC~=~OpenCode; OH~=~OpenHands. Dashes: not run. $^{\dagger}$Grok~4.7 runs only in its own harness, Grok Build.};
\end{tikzpicture}%
\arrayrulecolor{pleHairline}
\end{table}
\FloatBarrier

\section{Expert Seed Nomination Interface}
\label{app:nomination}

To ensure that every co-authored task in \bench{} originates from an authentic, economically valuable engineering problem rather than a synthetic prompt, practising physical engineers nominate seed tasks through our web-based \emph{Task Nomination Interface} (\cref{fig:nomination_ui}). Rather than writing a free-form prompt, the nominating engineer fills out six structured specification fields:
\begin{enumerate}[leftmargin=*,itemsep=2pt]
  \item \textbf{Situation:} Realistic engineering background, project context, and operational persona for the assignment.
  \item \textbf{Inputs:} High-level description of the starting project assets the engineer would expect in the workspace (e.g.\ CAD/BIM models, schedules, geotechnical or weather data).
  \item \textbf{Goal:} Quantifiable engineering target or economic optimisation objective (e.g.\ minimising installed material cost, peak load, or routing length).
  \item \textbf{Constraints:} Governing building codes, physical feasibility boundaries, safety limits, and hardware compatibility rules.
  \item \textbf{Verification:} High-level specification of how an automated deterministic script, simulator, or rubric should evaluate the submitted deliverable.
  \item \textbf{Expected Output:} Exact production deliverable formats required from the solver agent (e.g.\ XLSX schedules, DXF drawings, IFC4 BIM models).
\end{enumerate}
Once submitted, this structured nomination serves as the \emph{Expert Seed Prompt} (Stage~1 in \cref{fig:pipeline}) that guides the Task Creator agent's deep research, multi-format environment construction, and executable verifier synthesis.

\begin{figure*}[h!]
\centering
\definecolor{uibg}{HTML}{F5F7FA}\definecolor{uibdr}{HTML}{C8D2E0}%
\definecolor{uiink}{HTML}{0B192C}\definecolor{uimuted}{HTML}{5D6577}%
\definecolor{uiblue}{HTML}{1F2F86}\definecolor{uibluebg}{HTML}{E8ECF6}\definecolor{uibluebdr}{HTML}{B7BEDB}%
\definecolor{uired}{HTML}{E56B4E}\definecolor{uigreen}{HTML}{1F2F86}%
\definecolor{uiinput}{HTML}{F5F7FA}%
\resizebox{\textwidth}{!}{%
\begin{tikzpicture}[
  x=1.24cm, y=1.14cm,
  font=\sffamily,
  win/.style={draw=pleBorder, line width=0.95pt, fill=plePrint, rounded corners=5pt},
  fieldcard/.style={draw=pleHairline, line width=0.6pt, fill=pleSheet, rounded corners=3.5pt},
  textbox/.style={draw=pleHairline, line width=0.45pt, fill=plePrint, rounded corners=2pt},
  lbl/.style={font=\sffamily\scriptsize\bfseries, text=pleBlueline, anchor=west},
  desc/.style={font=\sffamily\tiny, text=pleGraphite, anchor=west},
  sample/.style={font=\sffamily\tiny, text=pleInk, anchor=north west, align=left},
]

\draw[win] (0,0) rectangle (16.4,8.55);

\fill[pleWash, rounded corners=5pt] (0,7.90) rectangle (16.4,8.55);
\fill[pleWash] (0,7.90) rectangle (16.4,8.20);
\draw[pleHairline, line width=0.65pt] (0,7.90) -- (16.4,7.90);
\fill[pleRedline]   (0.42,8.23) circle (2.4pt);
\fill[plePillBdr]   (0.74,8.23) circle (2.4pt);
\fill[pleBlueline]  (1.06,8.23) circle (2.4pt);
\node[font=\sffamily\scriptsize\bfseries, text=pleBlueline, anchor=west] at (1.45,8.23)
      {\bench{} TASK NOMINATION INTERFACE \ --- \ EXPERT SEED PROMPT};

\node[font=\sffamily\footnotesize\bfseries, text=pleInk, anchor=west] at (0.40,7.48)
      {PROPOSE NEW BENCHMARK TASK};
\node[font=\sffamily\tiny, text=pleGraphite, anchor=west] at (0.40,7.14)
      {Specify the six structured seed fields below (illustrated with the \emph{LV Cable Sizing} nomination):};

\draw[fieldcard] (0.40,5.04) rectangle (8.05,6.88);
\fill[pleWash, rounded corners=2pt] (0.52,6.46) rectangle (7.93,6.78);
\node[lbl] at (0.64,6.62) {1.\ SITUATION};
\node[desc] at (0.58,6.24) {Realistic engineering background, context, and operational persona:};
\draw[textbox] (0.58,5.18) rectangle (7.87,6.06);
\node[sample, text width=8.65cm] at (0.68,5.98)
      {``Act as the lead electrical engineer for a naval support facility. Size 32 low-voltage feeders and branch circuits (480\,V / 208\,V) across underground duct banks, cable trays, and indoor conduits to issue the permit cable schedule.''};

\draw[fieldcard] (8.35,5.04) rectangle (16.00,6.88);
\fill[pleWash, rounded corners=2pt] (8.47,6.46) rectangle (15.88,6.78);
\node[lbl] at (8.59,6.62) {2.\ INPUTS};
\node[desc] at (8.53,6.24) {Assets provided in the starting workspace (CAD/BIM, data, schedules):};
\draw[textbox] (8.53,5.18) rectangle (15.82,6.06);
\node[sample, text width=8.65cm] at (8.63,5.98)
      {``Single-line riser diagram (DXF), panel \& motor load schedules (XLSX), geotechnical soil thermal resistivity report (PDF), owner criteria with superseded revision, and vendor cable catalog \& price book.''};

\draw[fieldcard] (0.40,3.00) rectangle (8.05,4.84);
\fill[pleWash, rounded corners=2pt] (0.52,4.42) rectangle (7.93,4.74);
\node[lbl] at (0.64,4.58) {3.\ GOAL};
\node[desc] at (0.58,4.20) {Quantifiable engineering target or economic optimisation objective:};
\draw[textbox] (0.58,3.14) rectangle (7.87,4.02);
\node[sample, text width=8.65cm] at (0.68,3.94)
      {``Minimise total installed conductor and conduit material cost (\$) across all 32 circuits while satisfying all ampacity, thermal derating, voltage-drop, and equipment termination limits.''};

\draw[fieldcard] (8.35,3.00) rectangle (16.00,4.84);
\fill[pleWash, rounded corners=2pt] (8.47,4.42) rectangle (15.88,4.74);
\node[lbl] at (8.59,4.58) {4.\ CONSTRAINTS};
\node[desc] at (8.53,4.20) {Hard physical boundaries, safety rules, and governing building codes:};
\draw[textbox] (8.53,3.14) rectangle (15.82,4.02);
\node[sample, text width=8.65cm] at (8.63,3.94)
      {``NEC \& UFC 3-501-01 code limits, Neher--McGrath duct-bank mutual heating derating, ${\le}\,3\%$ feeder / ${\le}\,5\%$ total voltage drop, fire-pump locked-rotor drop, and max 5 cable sizes above \#1~AWG.''};

\draw[fieldcard] (0.40,0.96) rectangle (8.05,2.80);
\fill[pleWash, rounded corners=2pt] (0.52,2.38) rectangle (7.93,2.70);
\node[lbl] at (0.64,2.54) {5.\ VERIFICATION};
\node[desc] at (0.58,2.16) {Deterministic verification script, test suite, or simulation engine:};
\draw[textbox] (0.58,1.10) rectangle (7.87,1.98);
\node[sample, text width=8.65cm] at (0.68,1.90)
      {``Python verifier checking 25 hard \& 10 soft constraints per circuit, source-clause citation register accuracy, and installed package cost against the certified relaxation bound (\cref{eq:score}).''};

\draw[fieldcard] (8.35,0.96) rectangle (16.00,2.80);
\fill[pleWash, rounded corners=2pt] (8.47,2.38) rectangle (15.88,2.70);
\node[lbl] at (8.59,2.54) {6.\ EXPECTED OUTPUT};
\node[desc] at (8.53,2.16) {Exact production deliverable files required from the solver agent:};
\draw[textbox] (8.53,1.10) rectangle (15.82,1.98);
\node[sample, text width=8.65cm] at (8.63,1.90)
      {``Completed 32-circuit cable \& conduit schedule workbook (XLSX), updated riser plan sheet (DXF), and source-traced code requirement register (JSON).''};

\fill[pleBlueline, rounded corners=2.5pt] (12.80,0.22) rectangle (16.00,0.74);
\node[font=\sffamily\scriptsize\bfseries, text=white] at (14.40,0.48) {SUBMIT SEED PROMPT $\rightarrow$};

\end{tikzpicture}%
}
\caption{Demonstration diagram of the \bench{} \emph{Task Nomination Interface} used by practising physical engineers to author the \emph{Expert Seed Prompt} (Stage~1 of \cref{fig:pipeline}). Engineers specify the real-world engineering \textbf{Situation}, starting \textbf{Inputs}, quantifiable optimisation \textbf{Goal}, physical and code \textbf{Constraints}, deterministic \textbf{Verification} logic, and \textbf{Expected Output} deliverables (illustrated here with the \emph{LV Cable Sizing} task).}
\label{fig:nomination_ui}
\end{figure*}

\end{document}